\documentclass[pmlr,twocolumn,10pt]{jmlr}

\mlhtrack{proceedings}

\newif\iffinal
\finaltrue   % camera-ready

\iffinal
    \ifmlhneedspmlr
      \jmlrvolume{346}   % PMLR volume, given by the publication chairs 2026-11-06
      \jmlryear{2026}
    \fi
    \jmlrworkshop{Machine Learning for Health (ML4H) 2026}
\else
    \jmlrproceedings{}{Submitted to ML4H 2026: \mlhtrackname}
    \jmlrworkshop{Machine Learning for Health (ML4H) 2026}
\fi

\usepackage{booktabs}
\usepackage{csvsimple}
\usepackage{colortbl}
\definecolor{terrA}{HTML}{CFE3F5}
\definecolor{terrB}{HTML}{E9B77A}
\definecolor{uncov}{HTML}{EFEFEF}
\definecolor{noncl}{HTML}{DCDCDC}
\definecolor{markA}{HTML}{2C6FA6}
\definecolor{markB}{HTML}{B26A16}
\newcommand{\mka}{{\color{markA}\scriptsize$\leftarrow\!c_1$}}
\newcommand{\mkb}{{\color{markB}\scriptsize$\leftarrow\!c_2$}}
\newcommand{\ca}{\cellcolor{terrA}}
\newcommand{\cb}{\cellcolor{terrB}}
\newcommand{\cu}{\cellcolor{uncov}}
\newcommand{\cn}{\cellcolor{noncl}}
\newcommand{\swatch}[1]{\raisebox{-0.15ex}{\setlength{\fboxsep}{0.5pt}\fcolorbox{black!30}{#1}{\rule{0pt}{1ex}\hspace{1ex}}}}

\newcommand{\vendormark}[1]{\raisebox{-0.15ex}{\includegraphics[height=1.5ex]{image/logos/#1.pdf}}\,}
\newcommand{\vendorcell}[1]{\ifcsvstrcmp{#1}{}{\hphantom{\vendormark{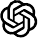}}}{\vendormark{#1}}}
\newcommand{\bestif}[2]{\ifcsvstrcmp{#1}{1}{\textbf{#2}}{#2}}
\newsavebox{\funnelbox}

\newcommand{\drop}[1]{{\scriptsize\textcolor{black!55}{($-$#1\%)}}}

\newcommand{\gatedsm}[1]{{\scriptsize\textcolor{black!62}{(#1)}}}
\usepackage[switch]{lineno}   % line numbers for review
\usepackage{comment}
\usepackage[breakable,skins]{tcolorbox}  % prompt boxes in the appendix
\usepackage{textcomp}         % straight quotes inside those boxes
\usepackage{listings}
\lstdefinestyle{prompt}{%
  basicstyle=\ttfamily\scriptsize,
  breaklines=true, breakatwhitespace=false, columns=fullflexible,
  breakindent=0pt,
  keepspaces=true, showstringspaces=false, extendedchars=true,
  aboveskip=0pt, belowskip=0pt,
  literate={—}{{---}}3 {–}{{--}}2 {…}{{...}}3 {≥}{{$\geq$}}1 {≤}{{$\leq$}}1
           {≠}{{$\neq$}}1 {→}{{$\rightarrow$}}1 {⇄}{{$\rightleftarrows$}}1
           {±}{{$\pm$}}1 {×}{{$\times$}}1 {÷}{{$\div$}}1 {≈}{{$\approx$}}1
           {∆}{{$\Delta$}}1 {'}{{\textquotesingle}}1 {`}{{\textasciigrave}}1%
}
\newtcolorbox{promptbox}[1]{breakable, colback=black!3, colframe=black!35,
  title=#1, fonttitle=\bfseries\footnotesize, boxrule=0.4pt,
  left=3pt, right=3pt, top=3pt, bottom=3pt, middle=2pt}
\tcbuselibrary{skins}
\usepackage{float}          % [H] placement, used in the appendix
\usepackage{shortcuts}        % project macros — provides \jsz{...} draft notes
\usepackage{capt-of}         % \captionof, so one float can carry a Table and a Figure
\usepackage{soul}             % \hl for the unbacked spans in the claim-support examples

\definecolor{unbackedtint}{RGB}{254,246,200}
\sethlcolor{unbackedtint}
\soulregister{\textless}{0}
\soulregister{\textgreater}{0}
\newcommand{\ub}[1]{\hl{#1}}

\title[Verifiable by Construction]{Verifiable by Construction: How Well Do LLMs Substantiate Their Clinical Claims with Verbatim Quotes?}

\iffinal
\author{%
  \Name{Jiashuo Zhang} \Email{jzhan427@jhu.edu}\\
  \addr Department of Computer Science, Johns Hopkins University
  \AND
  \Name{Yuling Chen} \Email{ychen408@jhu.edu}\\
  \addr School of Nursing, Johns Hopkins University
  \AND
  \Name{Yvonne Commodore-Mensah} \Email{ycommod1@jhmi.edu}\\
  \addr School of Nursing, Johns Hopkins University
  \AND
  \Name{Michael Oberst} \Email{moberst@jhu.edu}\\
  \addr Department of Computer Science, Johns Hopkins University
}
\else
\author{Anonymous Author(s)}
\linenumbers
\fi

\begin{document}

\maketitle

%%%%%%%%%%%%%%%%%%%%%%%%%%%%%%%%%%%%%%%%%%%%%%%%%%%%%%%%%%%%%%%%%%%%%%%%%%%%%%
%%  0.Abstract
%%%%%%%%%%%%%%%%%%%%%%%%%%%%%%%%%%%%%%%%%%%%%%%%%%%%%%%%%%%%%%%%%%%%%%%%%%%%%%
\begin{abstract}
Large language models (LLMs) have been widely adopted for clinical question answering (QA). Current systems can attach citations to their answers, but these often point to broad texts, leaving time-pressed clinicians unable to verify them efficiently. An alternative is to ensure that responses are verifiable by construction: providing fine-grained verbatim quotes from reference material that substantiate claims, so users can verify an answer without opening other documents. In this paper, we evaluate the ability of current models to perform this task end-to-end: from providing citations for every factual claim, to producing verbatim quotes, to ensuring that those quotes fully substantiate the claims. To do so, we build a standardized harness over four clinical practice guidelines and evaluate fourteen LLMs on 222 synthetic clinical questions, measuring each of these stages separately. We find that most models can attach verbatim quotes to over 90\% of their claims from prompting alone, apart from some lightweight models such as \texttt{claude-haiku-4.5}. Yet these quotes often fail to substantiate every detail of the claims they accompany. For instance, \texttt{claude-opus-5} produces verbatim quotes for 98.0\% of its claims, but fully substantiates only 37.1\%. Our work provides insights into the current capability gap of LLMs in building verifiable clinical QA systems, along with artifacts for future research.
\end{abstract}


% MANDATORY for ML4H — omitting these is grounds for DESK REJECTION.
\ifmlhneedsstatements
\paragraph*{Data and Code Availability} Data are from publicly available clinical guidelines issued by the American Heart Association and the American Diabetes Association. The code and the synthetic dataset are available at \url{https://github.com/oberst-lab/verifiable-by-construction}.
\paragraph*{Institutional Review Board (IRB)} IRB review was not required for this study. We use a synthetic dataset in this work, and no Protected Health Information (PHI) is involved.
\fi

%%%%%%%%%%%%%%%%%%%%%%%%%%%%%%%%%%%%%%%%%%%%%%%%%%%%%%%%%%%%%%%%%%%%%%%%%%%%%%
%%  1.Introduction
%%%%%%%%%%%%%%%%%%%%%%%%%%%%%%%%%%%%%%%%%%%%%%%%%%%%%%%%%%%%%%%%%%%%%%%%%%%%%%
\section{Introduction}
\label{sec:intro}

\newcommand{\overviewscale}{0.75}
\begin{figure*}[t]
\centering
\includegraphics[width=\overviewscale\textwidth]{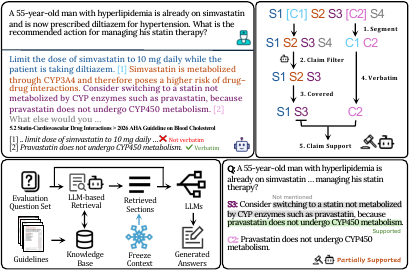}
\caption{
  \textbf{Our evaluation framework}:
  \textbf{Top left}: An example question and its answer with inline citations. A citation covers the sentence that precedes it.
  \textbf{Top right}: The pipeline of the automated evaluation. See Section~\ref{subsec:evaluation} for details.
  \textbf{Bottom left}: The dataflow of the answer generation process.
  \textbf{Bottom right}: An example of a claim support verdict made by an LLM judge.
}
\label{fig:overview}
\end{figure*}

Large language models (LLMs) have shown strong capabilities in natural language understanding and generation, and have been widely adopted across domains including healthcare~\citep{liu2025application}. In clinical question answering (QA), they assist healthcare professionals by producing evidence-based responses to complex medical queries~\citep{zakka2024almanac, kresevic2024optimization, ong2024surgeryllm, ke2025retrieval}. Though LLMs produce fluent and contextually relevant outputs, they sometimes hallucinate, i.e., state information that is factually incorrect~\citep{omar2025multi, yoon2025navigating}. In high-stakes fields like healthcare, where such errors can have serious consequences, professionals may not have time to verify LLM outputs against cited sources.

With this challenge in mind, evaluation of clinical QA systems has extended beyond the quality of the answer itself, to include measures of how ``easy to verify'' an answer is~\citep{slobodkin2024attribute, feng2026expert}. A common approach to help users verify an answer is to attach citations to the output~\citep{liu2023evaluating, gao2023enabling}, a feature that clinical QA systems have now incorporated~\citep{wu2025automated, wang2025medcite}. However, citations typically point to broad texts, such as webpages, entire papers~\citep{nakano2021webgpt, bohnet2022attributed}, or sections within guidelines~\citep{carl2025enhancing}. As a result, a citation itself is not sufficient to verify an answer, but may require readers to track down which part of the cited document supports a claim~\citep{zhao2024awecita}. Moreover, readers often equate citations with credible proof, so the mere presence of citations can create a false sense of validation~\citep{venkit2024search}.

Finer attribution is a way to address this problem, by providing specific supporting snippets of text from sources instead of broad citations.  Ideally, this approach allows for faster and more precise assessment of support~\citep{slobodkin2024attribute, hennigen2023towards}, without the need to locate the evidence in external documents.

In this paper, we evaluate current models on this task, using an end-to-end automated evaluation framework to simulate the verification process a user might follow (Figure~\ref{fig:overview}). First, we set up a retrieval-based clinical QA system using a set of clinical guidelines and instructions to answer questions with in-line quotations of supporting evidence. Using a dataset of synthetic clinical queries, we then examine each answer one stage at a time: filtering for factual claims, and then assessing whether every claim has a citation, whether the quotations provided are present (verbatim) in the source, and whether these quotes on their own fully substantiate the claim.

We report our findings in Section~\ref{sec:results}, where we observe that no model consistently achieves this goal, with the best-performing systems (e.g., \texttt{gpt-5.4}) supporting around 75\% of claims with exact substantiating evidence.  More notably, our framework allows us to decompose performance across stages, and observe that the relatively poor performance of some models (e.g., \texttt{claude-haiku-4.5}, which achieves a rate of 24.9\%) is driven in large part by a failure to produce faithful verbatim quotations, while other models (e.g., \texttt{claude-opus-5}) produce verbatim quotations, but fail to provide enough context to fully substantiate their claims. Our primary contributions are:
\begin{itemize}
  \item We develop an end-to-end evaluation framework that probes verifiability of answers by checking claims against their supporting quotes from cited sources, including deterministic checks (e.g., to ensure quotes are verbatim) as well as human-validated LLM judges (e.g., to assess whether or not quotes fully support claims).
  \item Using this framework, we compare fourteen mainstream LLMs from six vendors on a dataset of 222 synthetic clinical queries designed to be answerable from a set of four clinical guidelines, and derive insights into drivers of relative model performance.
  \item We release our question-answering framework with the accompanying harness and our synthetic question set described above.
\end{itemize}

%%%%%%%%%%%%%%%%%%%%%%%%%%%%%%%%%%%%%%%%%%%%%%%%%%%%%%%%%%%%%%%%%%%%%%%%%%%%%%
%%  2.Related Work
%%%%%%%%%%%%%%%%%%%%%%%%%%%%%%%%%%%%%%%%%%%%%%%%%%%%%%%%%%%%%%%%%%%%%%%%%%%%%%
\section{Related Work}
\label{sec:related}
\paragraph{Attributed answer generation.}
Open-domain question-answering systems attribute an answer to the documents or passages it draws on~\citep{nakano2021webgpt,bohnet2022attributed}, and models can be trained to cite those passages more faithfully~\citep{huang2024training}. A finer grain of attribution can tie each claim to a verbatim supporting quote~\citep{menick2022teaching} or a concise evidence span~\citep{slobodkin2024attribute}.
\citet{zhang2025longcite} propose LongCite, fine-tuning long-context models to emit sentence-level citations that index the supporting sentences in the provided context. \citet{shao2025dr} introduce DR~Tulu, a deep-research agent that produces long-form answers with snippet-level citations.

\paragraph{Evaluation frameworks.}
Evaluation of clinical QA systems has moved from multiple-choice licensing exams toward open-ended, clinician-grounded assessment. MedHELM~\citep{bedi2026holistic} scores models across a clinician-validated taxonomy of real-world clinical tasks with a jury of LLM judges. HealthBench~\citep{arora2025healthbench} grades multi-turn health conversations against physician-written rubrics. These benchmarks often judge whether answers are correct, complete, and grounded, but do not directly assess the ``verifiability'' of answers. In the context of natural language generation, \citet{rashkin2023measuring} propose a framework for judging whether a statement is supported by its cited source, and \citet{gao2023enabling} automate a similar process with a natural-language-inference (NLI) model.

Closer to our work, SourceCheckup~\citep{wu2025automated} audits the web sources an LLM cites for a medical question and finds, with an LLM judge, that half or more of its responses are not fully supported. MedCite~\citep{wang2025medcite} builds a PubMed retrieval pipeline and scores citation recall and precision with an LLM attribution judge. VERICITE~\citep{ma2026vericite} checks each citation at the sentence level with an NLI model, reporting that only a minority of cited claims are entailed by their source.  However, these prior works focus primarily on whether or not a cited document supports claims, while we focus more specifically on verifiability, which includes (a) assessing the density of support (i.e., whether or not every claim has an accompanying quotation), (b) checking whether extracted quotations are correctly reproduced, and (c) assessing whether or not these quotations are sufficient to fully substantiate each claim.

%%%%%%%%%%%%%%%%%%%%%%%%%%%%%%%%%%%%%%%%%%%%%%%%%%%%%%%%%%%%%%%%%%%%%%%%%%%%%%
%%  3.Methods
%%%%%%%%%%%%%%%%%%%%%%%%%%%%%%%%%%%%%%%%%%%%%%%%%%%%%%%%%%%%%%%%%%%%%%%%%%%%%%
\section{Methods}
\label{sec:methods}

We ground our evaluation in a concrete task.  Motivated by applications in cardiovascular risk management, we build a clinical QA system\footnote{An illustrative user interface is shown in Appendix~\ref{app:system-ui}, though we stress that our focus in this work is on evaluating the narrower property of verifiability, rather than evaluating the performance of the system as a whole.} for answering questions regarding practice guidelines.  Such a system could be useful for care team members who engage in shared decision-making conversations. Given a question, this system retrieves and reads relevant sections from a curated corpus of practice guidelines, and is designed to generate answers in which every claim is accompanied by an inline citation that names the source section and reproduces its exact wording.

In Section~\ref{subsec:system-design}, we describe this task (and clinical QA system) in more detail, and in Section~\ref{subsec:evaluation} we describe our evaluation framework and metrics, which we apply to the outputs of this system.

\subsection{Task and System Design}
\label{subsec:system-design}

\label{subsec:guideline}
\textbf{Guideline Corpus}.
To develop our task, we start from a corpus of four clinical practice guidelines from the American Heart Association (AHA) and the American Diabetes Association (ADA): the 2019 AHA guideline on the primary prevention of cardiovascular disease~\citep{doi:10.1161/CIR.0000000000000678}, the 2025 AHA guideline on high blood pressure~\citep{doi:10.1161/CIR.0000000000001356}, the 2026 AHA guideline on blood cholesterol~\citep{doi:10.1161/CIR.0000000000001423}, and the 2026 ADA Standards of Care in Diabetes~\citep{10.2337/dc26-SINT}, which together cover the principal modifiable risk factors for atherosclerotic cardiovascular disease.  Appendix~\ref{app:corpus} provides more detail on how this corpus is preprocessed for later retrieval.

\label{subsec:dataset}
\noindent\textbf{Synthetic Clinical Queries}.
From this corpus, we develop a set of synthetic clinical queries. Using example questions provided by clinical professionals for reference, we prompt \texttt{gpt-4o-mini} to synthesize 222 clinical questions in total (see Appendix~\ref{app:question-set} for details and Appendix~\ref{app:prompts-data} for the prompts) based on content present in the guidelines.  Notably, each of our clinical questions is derived from a particular subsection of a particular guideline, which serves two purposes:  first, it helps ensure that questions are reasonably answerable from the provided reference material, and second, it allows us to check that our retrieval process is of sufficiently high quality and does not confound our later evaluation (e.g., by failing to retrieve relevant material).

\label{subsec:semantic-retrieval}
\noindent\textbf{Clinical QA System}. Our system consists of two components.

\textit{LLM-guided semantic retrieval}:
Unlike a conventional vector index ranked by cosine similarity or a lexical method, our retrieval is driven directly by an LLM. Given the user's query, it selects the most relevant sections based on the section summaries in our knowledge base.  Because we know that the source subsection is a ``correct'' reference for answering a given query, we are able to evaluate and benchmark different approaches to performing retrieval. We find (in Appendix~\ref{subsec:retrieval-results}) that LLM-guided semantic retrieval substantially outperforms both a dense-vector baseline (\texttt{text-embedding-3-small} from OpenAI) and a lexical baseline (Okapi BM25), returning fewer sections whose set contains the ground-truth section more often. The implementation details are provided in Appendix~\ref{app:prompts-retrieval}.  We then fix the retrieval stage, so that retrieval quality does not confound the comparison between generation models. We select \texttt{deepseek-v4-flash} for a one-time retrieval, balancing retrieval quality against cost-efficiency, and freeze its output as the shared context for all subsequent generation models. Per-model generation configurations are provided in Appendix~\ref{app:gen-config}.

\textit{Question answering with verbatim quotes}:
\label{subsec:verbatim-citation}
Given the query and the provided retrieved context, the system prompt (Appendix~\ref{app:prompts-generation}) instructs the model to answer in short prose and to attach an inline citation marker to each clinical claim. A marker pairs the identifier of one retrieved section with a short quote copied verbatim from that section, and is written as \texttt{\{\{cite:\textit{guideline}:\textit{section}|\textit{quote}\}\}}.  We do so to mimic the design of a system that could then render these markers in a user interface as numeric citation indicators (e.g., $[1], [2]$) that link to the corresponding quotes. The model writes each claim as a stand-alone sentence before its marker is attached, so the answer remains complete clinical prose once every marker is removed. Therefore, a claim and its quote form a self-contained unit, and the two can be processed as separate objects (as shown in the top-right panel of Figure~\ref{fig:overview}).

\subsection{Evaluation Framework and Metrics}
\label{subsec:evaluation}

Our evaluation framework proceeds in stages:

\noindent\textbf{Claim Filter}.  We define a \emph{claim sentence} as any sentence that states at least one checkable clinical claim. Ideally, every clinical claim is grounded in the context and carries a citation that supports it.  To identify claims, we first strip citation markers from the full answer, recording positions, and split remaining prose into sentences using a deterministic segmenter. A \emph{claim filter} (driven by \texttt{deepseek-v4-flash}) then identifies sentences that contain claims, which are retained for further analysis. Implementation details are provided in Appendix~\ref{app:claim-units}.

\noindent\textbf{Citation Coverage}.  We consider a claim to be ``covered'' by a citation using an  \emph{attribution window} of size $k$, where a citation is considered to cover the $k$ claim sentences immediately preceding it, but never reaching past the previous citation (see Appendix~\ref{app:window}). For the remainder of the main text, we use $k = 1$ throughout, where a citation applies only to the nearest claim sentence, but we assess sensitivity to this choice in Appendix~\ref{app:coverage-full}. \textit{Citation Coverage} (CC) is then defined as the fraction of claim sentences that are covered by at least one citation.
\label{subsec:coverage}

\noindent\textbf{Verbatim Compliance}.
\label{subsec:vcr}
As described in Section~\ref{subsec:verbatim-citation}, the system prompt requires every citation to quote its source verbatim, but an instruction is not a guarantee. In practice, a quote sometimes departs from the source. We consider a quotation ``verbatim compliant'' if it appears in the cited section, and is either an \textit{exact} character-for-character match or satisfies one of two relaxed conditions: \emph{normalized} and \emph{elided}. The \emph{normalized} category allows for case inconsistencies, minor symbol differences, and other formatting variation, while the \emph{elided} category permits an ellipsis in the middle of the quote. The detailed criteria are described in Appendix~\ref{app:match-tiers}.  We then define the \textit{Verbatim Compliance Rate} (VCR) as the fraction of those citations whose quote passes the check.

\label{subsec:claim-support}
\noindent\textbf{Claim Support}.
We consider a claim to be supported if accompanying citations (or quotes) that cover it are enough to substantiate the claim, so that a reader can verify the claim on the spot. Each claim sentence is paired with the quotes cited for it under the attribution window $k = 1$, and an LLM judge decides how much of the claim those quotes support. The judge is given the question, the claim sentence, and its quotes, and nothing else. The verdict is one of four categories: \emph{fully supported} (the quotes support every point of the claim), \emph{partially supported} (the quotes address some points of the claim but not all), \emph{not supported} (the quotes do not support any point of the claim), or \emph{contradicted} (the quotes conflict with the claim). The full prompt is provided in Appendix~\ref{app:prompts-claim-support}.  The \textit{Claim Support Rate} (CSR) considers only those claim sentences with at least one verbatim quotation, and is provided in two forms:  The \textit{strict} variant is the fraction of those claim sentences where the verbatim quote(s) fully support the claim, and the \textit{lenient} variant is the fraction where the verbatim quote(s) are deemed to at least \textit{partially support} the claim.

\label{subsec:claim-funnel}
\noindent\textbf{The Claim Funnel}.  While the preceding metrics each score one property of an answer (see Figure~\ref{fig:claim-funnel} for an overview), the \emph{claim funnel} sequentially passes every claim sentence through those checks and records the number and proportion of claim sentences that remain after each stage. As shown in the top-right of Figure~\ref{fig:overview}, each claim sentence is checked for a covering citation, then for a quote that passes the verbatim check, and finally judged for claim support. We define the \emph{certified claim rate} (CCR) as the share of all claim sentences that carry at least one verbatim-compliant quote and are fully supported by the quote(s) that remain.

\newcommand{\funnelgraphscale}{1.0}
\begin{figure}[t]
\centering
\includegraphics[width=\funnelgraphscale\columnwidth]{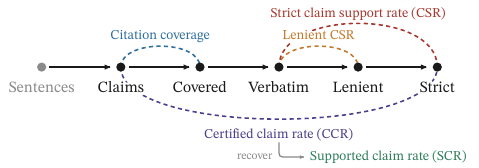}
\caption{The visualized relationships between stages and metrics.}
\label{fig:claim-funnel}
\end{figure}

%%%%%%%%%%%%%%%%%%%%%%%%%%%%%%%%%%%%%%%%%%%%%%%%%%%%%%%%%%%%%%%%%%%%%%%%%%%%%%
%%  4.Results
%%%%%%%%%%%%%%%%%%%%%%%%%%%%%%%%%%%%%%%%%%%%%%%%%%%%%%%%%%%%%%%%%%%%%%%%%%%%%%
\section{Results}
\label{sec:results}

\newcommand{\funnelscale}{0.85}
\newcommand{\stagesscale}{0.85}
\begin{table*}[htbp]
\centering
\makeatletter
\long\def\@makecaption#1#2{\vskip10pt{\raggedright #1: #2\par}\vskip7pt}
\makeatother
\caption{The claim funnel. The four stage columns are percentages of that model's claims, so a row reads left to right as one ``funnel'' with a fixed denominator. The final columns (Recov. and SCR) are described further in Section~\ref{subsec:recovery-results}. Percentages in parentheses show the relative loss against the column to its left. \emph{Sent.}, \emph{Claims} and \emph{Recov.} are counts. Appendix~\ref{app:round-variance} gives the three-round range.}
\label{tab:claim-funnel}
\sbox{\funnelbox}{%
{\small
\setlength{\tabcolsep}{2.3pt}
\begin{tabular}{@{}l@{\hspace{8pt}}lllllc@{\hspace{7pt}}|@{\hspace{7pt}}lc@{}}
\toprule
& & & \multicolumn{4}{c@{\hspace{7pt}}@{\hspace{7.4pt}}}{Claims remaining after each stage (\%)} & & (\%) \\
\cmidrule(lr{15.2pt}){4-7}
Model & Sent. & Claims & Covered & Verbatim & Lenient
  & \multicolumn{1}{c@{\hspace{7pt}}@{\hspace{7.4pt}}}{CCR\,{\scriptsize(Strict)}} & Recov. & SCR \\
\midrule
\csvreader[late after line=\\, filter strcmp={\colmain}{1}]%
  {tables/claim_funnel.csv}{maintable=\colmain, vendor=\colvendor, model=\colmodel,
   sentences=\colsent, claims=\colclaims,
   pcovered=\colpcov, pverbatim=\colpverb, plenient=\colpleni,
   recovered=\colrecov, ccr=\colccr, scr=\colscr,
   dropclaim=\coldclaim, dropcited=\coldcited, dropverbatim=\coldverb,
   droplenient=\coldleni, dropcert=\coldcert,
   bestccr=\colbccr, bestscr=\colbscr}%
  {\vendorcell{\colvendor}\texttt{\colmodel} & \colsent
   & \colclaims & \colpcov\,\drop{\coldcited}
   & \colpverb\,\drop{\coldverb} & \colpleni\,\drop{\coldleni}
   & \bestif{\colbccr}{\colccr}\,\drop{\coldcert}
   & \colrecov & \bestif{\colbscr}{\colscr}}
\bottomrule
\end{tabular}}
}%
\scalebox{\funnelscale}{\usebox{\funnelbox}}

\vspace{9pt}
\includegraphics[width=\stagesscale\textwidth]{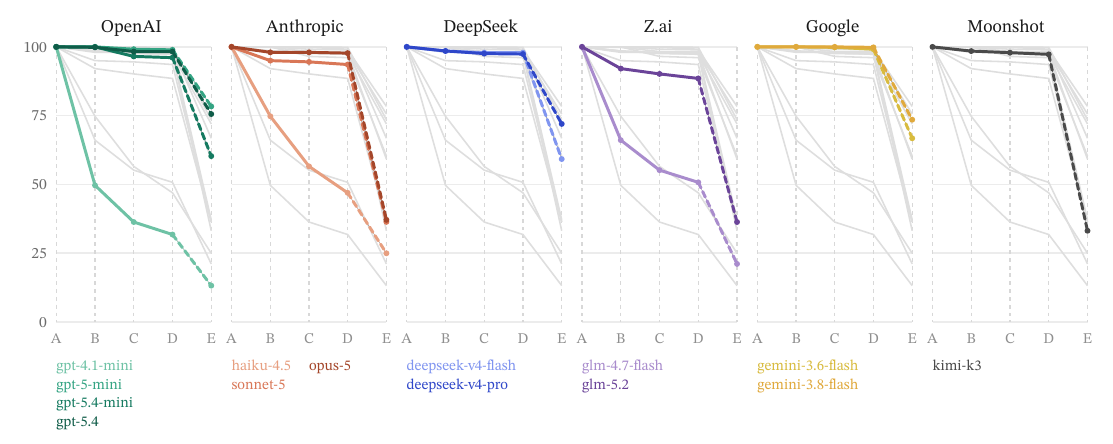}
\captionof{figure}{The claim funnel per provider. Stages: A claims, B covered,
C verbatim, D lenient, E strict.}
\label{fig:claim-funnel-stages}
\end{table*}

Our full results of the claim funnel analysis are shown in Table~\ref{tab:claim-funnel} and Figure~\ref{fig:claim-funnel-stages}.  No model perfectly completes the task, and there are clear differences in performance across models.  For instance, the highest observed CCRs are 72--78\%, from larger-scale or more recent models (e.g., \texttt{deepseek-v4-pro}, \texttt{gpt-5-mini}, \texttt{gpt-5.4}), leaving room for improvement.  Moreover, end-to-end performance separates larger-scale models (e.g., \texttt{gpt-5.4}) and smaller-scale models (e.g., \texttt{gpt-4.1-mini}), the former achieving a CCR of 75.6\% and the latter only 13.2\%.

Beyond these high-level results, the structure of our analysis allows us to decompose these rates into failures at distinct parts of the pipeline, leading us to a few additional findings.

First, the majority of models (with notable exceptions of \texttt{gpt-4.1-mini}, \texttt{claude-haiku-4.5}, and \texttt{glm-4.7-flash}) manage to produce sufficient verbatim quotes to cover a large fraction of their claims with potential supporting evidence (all over 90\%).  Those three smaller-scale models have substantial drop-offs, either in citation coverage (e.g., 66.1\% for \texttt{glm-4.7-flash}) or in verbatim compliance (e.g., 69.0\% of its citations for \texttt{claude-haiku-4.5}).

Second, many models have near-perfect performance at producing verbatim quotes that cover their claims, but fail to produce enough evidence to fully substantiate them.  For instance, 98\% of claims from \texttt{claude-opus-5} are accompanied by at least one verified verbatim quote, but only 37.1\% are deemed fully supported by their accompanying quotes.  Moreover, this drop-off is largely due to partial support, as opposed to a complete lack of evidence. We give examples of what the quotes leave unsupported, classify them into failure modes, and demonstrate that claim length plays a role but does not fully explain the gap.

Finally, we demonstrate that a substantial amount of performance is recoverable using the provided reference material, by searching the sections read by each model for fresh quotations that fully support the given claims. The share of remaining claims from \texttt{claude-opus-5} rises from 37.1\% to 85.1\% after recovery, while lightweight models like \texttt{gpt-4.1-mini} remain poor due to severe degradation in prior stages.

\subsection*{Finding 1: Small Models Fail the Citation Task}
\label{subsec:coverage-verbatim-results}

\newcommand{\citationqualityscale}{0.80}
\begin{table*}[t]
\centering
\caption{Citation coverage ($k = 1$) and verbatim compliance. \emph{Claims}, the number of sentences that passed the claim filter, is the denominator of both coverage columns; \emph{Cites}, the number of citations the model emitted, is the denominator of every compliance figure. The last column recomputes coverage over only the citations that pass the verbatim check. Appendix~\ref{app:coverage-full} gives the full coverage results.}
\label{tab:citation-quality}
\scalebox{\citationqualityscale}{{\small
\setlength{\tabcolsep}{4pt}
\begin{tabular}{@{}lccc@{\hspace{6pt}}|@{\hspace{6pt}}cccc@{\hspace{6pt}}|@{\hspace{6pt}}c@{}}
\toprule
& \multicolumn{3}{c@{\hspace{6pt}}@{\hspace{6.4pt}}}{Citation Coverage}
  & \multicolumn{4}{c@{\hspace{6pt}}@{\hspace{6.4pt}}}{Verbatim Compliance (\%)}
  & \multicolumn{1}{c}{Compliant} \\
\cmidrule(lr){2-4} \cmidrule(lr){5-8} \cmidrule(l){9-9}
Model & Claims & Cites & \multicolumn{1}{c@{\hspace{6pt}}@{\hspace{6.4pt}}}{Cov. (\%)}
  & Exact & Norm. & Elid. & \multicolumn{1}{c@{\hspace{6pt}}@{\hspace{6.4pt}}}{VCR}
  & Cov. (\%) \\
\midrule
\csvreader[late after line=\\, filter strcmp={\colmain}{1}]%
  {tables/citation_quality.csv}{vendor=\colvendor, model=\colmodel, maintable=\colmain,
   claims=\colclaims, cites=\colcites, cc1=\ccone, bestcc1=\bone,
   exact=\colexact, normalized=\colnorm, elided=\colelid,
   vcr=\colvcr, bestvcr=\colbvcr, cc1c=\ccgate, bestcc1c=\bgate}%
  {\vendorcell{\colvendor}\texttt{\colmodel} & \colclaims & \colcites
   & \bestif{\bone}{\ccone}
   & \colexact & \colnorm & \colelid & \bestif{\colbvcr}{\colvcr}
   & \bestif{\bgate}{\ccgate}}
\bottomrule
\end{tabular}}}
\end{table*}

Three lightweight models (\texttt{claude-haiku-4.5}, \texttt{gpt-4.1-mini} and \texttt{glm-4.7-flash}) fail to attach a verbatim quote to over 40\% of their claims (Table~\ref{tab:claim-funnel}). To understand what drives these results, we provide a more detailed breakdown of citations, coverage, and verbatim compliance, in Table~\ref{tab:citation-quality}. Apart from the three underperforming models, the remaining models show high citation coverage and verbatim compliance, with the bigger models generally outperforming the smaller ones. For instance, \texttt{claude-opus-5} produces the most citations and achieves the highest VCR at 100.0\%, outperforming \texttt{claude-sonnet-5}'s 98.9\%, while \texttt{claude-haiku-4.5} scores only 69.0\% even with the largest share of normalized matches. The failure to quote verbatim further lowers the coverage rate.

To build a qualitative understanding of these failures, we reviewed a sample of responses from each of the low-performing models, and present examples in Appendix~\ref{app:coverage-failures}. Anecdotally, we observe that models sometimes provide no citations (e.g., \texttt{glm-4.7-flash}). Another failure mode observed in \texttt{claude-haiku-4.5} is that it writes a summary or recap without supporting citations, which drags its coverage down. \texttt{gpt-4.1-mini} tends to place citations at the end of the answer. We also observe that \texttt{claude-haiku-4.5} frequently embeds quoted evidence directly into its responses, leaving little substantive claim content after citation marker removal and causing many sentences to fail the claim filter. All of these behaviors reflect incomplete adherence to the citation style specified in the system prompt.

\subsection*{Finding 2: Models struggle to provide strict supporting evidence}
\label{subsec:finding-support}

\newcommand{\claimsupportscale}{0.80}
\begin{table*}[htbp]
\centering
\caption{Claim support (as judged by \texttt{deepseek-v4-flash}). \emph{Cited} is the number of claims accompanied by at least one verbatim-compliant quote. \emph{Strict} gives the fraction of cited claims that are ``fully supported''; \emph{Lenient} also includes claims that are ``partially supported''. The \emph{Shortfall} counts sum to \emph{Partial} plus \emph{None}.}
\label{tab:claimsupport}
\scalebox{\claimsupportscale}{%
{\small
\setlength{\tabcolsep}{3pt}
\begin{tabular}{@{}l|c|cccc|cccc|cc@{}}
\toprule
\multicolumn{1}{@{}l}{} & \multicolumn{1}{c}{}
  & \multicolumn{4}{c}{Shortfall ($n$)}
  & \multicolumn{4}{c}{Verdict ($n$)}
  & \multicolumn{2}{c}{Rate (\%)} \\
\cmidrule(lr){3-6} \cmidrule(lr){7-10} \cmidrule(l){11-12}
\multicolumn{1}{@{}l}{Model} & \multicolumn{1}{c}{Cited}
  & Add. & Causal & Scope & \multicolumn{1}{c}{Strength}
  & Full & Partial & None & \multicolumn{1}{c}{Contra}
  & Strict & Lenient \\
\midrule
\csvreader[late after line=\\, filter strcmp={\colmain}{1}]%
  {tables/claim_support_gated.csv}{maintable=\colmain, vendor=\colvendor, model=\colmodel,
   paired=\colpaired, beststrict=\bstrict, bestlenient=\blenient,
   strict=\colstrict, lenient=\collenient,
   nfully=\colfull, npartial=\colpart, nnot=\colnot, ncontra=\colcontra,
   naddition=\coladd, ncausal=\colcausal, nscope=\colscope, nstrength=\colstrength}%
  {\vendorcell{\colvendor}\texttt{\colmodel} & \colpaired
   & \coladd & \colcausal & \colscope & \colstrength
   & \colfull & \colpart & \colnot & \colcontra
   & \bestif{\bstrict}{\colstrict} & \bestif{\blenient}{\collenient}}
\bottomrule
\end{tabular}}%
}
\end{table*}

Even for models that reliably include verbatim quotes, we observe substantial underperformance at the task of fully substantiating their claims. The full breakdown of claim support is judged by \texttt{deepseek-v4-flash}\footnote{Each verdict is the majority of three sampled judgements, with additional rounds run to break ties. }, as shown in Table~\ref{tab:claimsupport}. 
For most models, the cited quotes provide at least partial support for the vast majority of claims, as reflected in the high lenient support rates.  However, the strict claim support rates\footnote{The strict CSR is computed over \emph{cited} claims. The CCR has the same numerator, but over \emph{all} claims.} exhibit considerable variation, ranging from 33.8\% to 79.0\%. Cases where the quotes do not support the claim at all or even contradict it are rare across all models.

Given that the results depend on an LLM-as-judge, we validated it against two human annotators.\footnote{Both are authors of this paper.}  Each annotator independently labelled a stratified sample of 70 cases, blinded to the LLM judge labels.  Between the two human annotators, a high degree of agreement was observed (Cohen's $\kappa = 0.853$).  We further observed high rates of agreement between human annotators and the \texttt{deepseek-v4-flash} judgements, with Cohen's $\kappa$ of 0.858 and 0.810 for the two annotators.  We provide the full confusion matrices in Appendix~\ref{app:human-annotation}.
We also assessed agreement between \texttt{deepseek-v4-flash} and alternative judges (\texttt{gpt-5-mini}, \texttt{grok-4.3}) on full samples across 14 evaluated models, with Cohen's $\kappa$ of 0.788 and 0.647 indicating substantial agreement.  Results for alternative judges are shown in Appendix~\ref{app:claim-support-second}.

\newcommand{\examplesize}{\footnotesize}
\begin{figure}[t]
\centering
\examplesize
\begin{tcolorbox}[enhanced, colback=white, colframe=black!30,
  boxrule=0.4pt, left=5pt, right=5pt, top=3pt, bottom=3pt,
  title={\textsc{Unsupported addition} \hfill \normalfont\texttt{claude-opus-5}},
  coltitle=black, colbacktitle=black!6, fonttitle=\bfseries\examplesize]
\textbf{Claim.} She should also be counseled about low-dose aspirin 81 mg daily, \ub{started from 12 weeks' gestation}, to reduce her risk of preeclampsia.

\vspace{1pt}
\textbf{Quote.} \textit{should be counseled about the benefits of low-dose (81 mg/day) aspirin to reduce the risk of preeclampsia and its sequelae}
\end{tcolorbox}
\vspace{-3pt}
\begin{tcolorbox}[enhanced, colback=white, colframe=black!30,
  boxrule=0.4pt, left=5pt, right=5pt, top=3pt, bottom=3pt,
  title={\textsc{Unsupported addition} \hfill \normalfont\texttt{glm-5.2}},
  coltitle=black, colbacktitle=black!6, fonttitle=\bfseries\examplesize]
\textbf{Claim.} High-intensity statins \ub{(atorvastatin 40--80 mg or rosuvastatin 20--40 mg)} are preferred for patients at high ASCVD risk, as they can achieve \ub{$\geq$50\%} LDL-C reductions.

\vspace{1pt}
\textbf{Quote.} \textit{Atorvastatin and rosuvastatin can achieve high-intensity LDL-C reductions and are preferred for patients at high/very high ASCVD risk}
\end{tcolorbox}

\caption{Example of two partially supported claims exhibiting the \emph{unsupported addition} failure mode. Yellow highlights indicate content not grounded in the quote.}
\label{fig:support-main-examples}
\end{figure}

\newcommand{\supportlengthscale}{0.9}
\begin{figure}[t]
\centering
\includegraphics[width=\supportlengthscale\columnwidth]{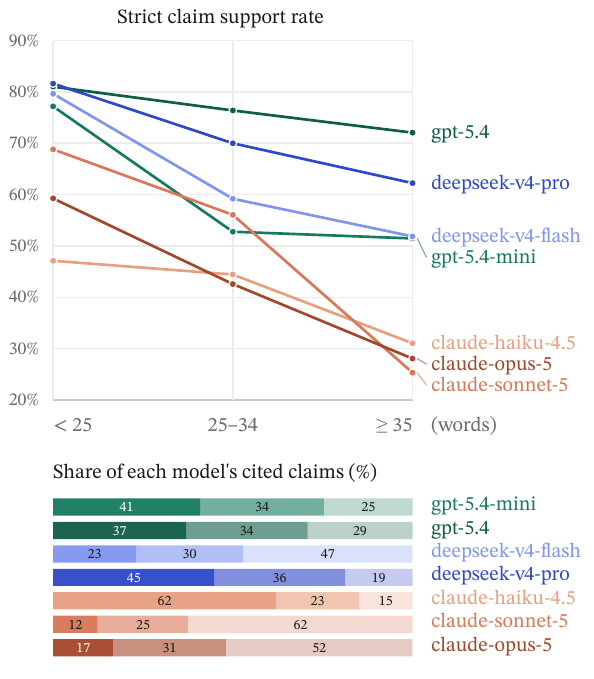}
\caption{Strict claim support rate against the length of the claim being judged, for selected models from OpenAI, Anthropic and DeepSeek. See Appendix~\ref{app:support-length} for the full results.}
\label{fig:support-length}
\end{figure}

We defined a taxonomy of four failure modes for partially supported and unsupported claims, drawing both on our own impressions and on existing schemes \citep{pagnoni2021understanding, yue2023automatic}:
\emph{Unsupported addition} covers cases where the quote fully substantiates some, but not all, of the statements in the claim.  \emph{Strength inflation} and \emph{scope inflation} cover cases where the strength of a recommendation (e.g., ``should'' versus ``must'') is inflated, or where the relevant population is inflated (e.g., a claim that applies to diabetic patients is applied to all patients). \emph{Causal upgrade} refers to the statement of an association as a cause. Figure~\ref{fig:support-main-examples} gives two observed examples, and Appendix~\ref{app:prompts-claim-support} the full definitions.

The vast majority of claims fall in the \emph{unsupported addition} category.  Given that longer claim sentences inherently require more citation support, we performed a sensitivity analysis in Figure~\ref{fig:support-length}, where we examined the rate of partial support as a function of claim length.  Longer claims do carry lower strict claim support rates, and \texttt{claude-sonnet-5} and \texttt{claude-opus-5} write substantially more sentences over 35 words than \texttt{gpt-5.4}. Even so, both Claude models sit below \texttt{gpt-5.4} within every one of the three length strata.  Hence, while the gap in performance may be partially explained by differences in length, this does not explain the full gap.

\subsection*{Finding 3: Most missing evidence is already in what models read}
\label{subsec:recovery-results}

With the hypothesis that models do compose the answers based on the reference, but the ability and style to attach the quotes may vary, we employ an LLM judge (\texttt{deepseek-v4-flash}) to recover claims that are partially supported due to the unsupported additions shortfall, by searching for missing evidence in the sections the model read. 
The judge takes one question, one claim and the sections the model read as input, and returns the evidence that fully substantiates the claim. The implementation details are in Appendix~\ref{app:claim-recovery}. We introduce the \emph{supported claim rate} (SCR) as the fraction of claims that are either fully supported or recovered from partially supported. The results are in the last column of Table~\ref{tab:claim-funnel}.

SCR is substantially higher than CCR across all models. \texttt{gpt-5-mini}, \texttt{gpt-5.4} and \texttt{gpt-5.4-mini} reach the highest supported claim rates, above 92.5\%. \texttt{claude-opus-5} rises from a CCR of 37.1\% to an SCR of 85.1\%, recovering most of its partially supported claims. However, for lightweight models like \texttt{claude-haiku-4.5}, SCR remains poor, given the ceiling of their already low lenient support rates. This indicates that many claims are grounded in the sections the model read, but the models fail to attach enough or appropriate quotes that establish them.

%%%%%%%%%%%%%%%%%%%%%%%%%%%%%%%%%%%%%%%%%%%%%%%%%%%%%%%%%%%%%%%%%%%%%%%%%%%%%%
%%  5.Discussion
%%%%%%%%%%%%%%%%%%%%%%%%%%%%%%%%%%%%%%%%%%%%%%%%%%%%%%%%%%%%%%%%%%%%%%%%%%%%%%
\section{Discussion}
\label{sec:discussion}

We build a clinical QA system that provides verbatim quotes from cited sources, and propose an end-to-end automated evaluation framework to simulate a human reader verifying the claims in an answer against the cited evidence. Our comparison across fourteen mainstream LLMs from six providers shows that most models can provide verbatim-compliant quotes from prompting alone, but the quality of the quotes and their support for the claims vary widely.

One limitation of our evaluation is that we use a synthetic question set, which may not fully reflect real-world clinical queries. To mitigate this, we referred to example clinical questions provided by professionals and conducted a small-scale annotation study to validate the reasonableness of the questions. The main reason for using synthetic questions is that our evaluation primarily focuses on models' generation capabilities. We therefore wanted a set of questions for which the relevant answers are known to be present in the source materials, and synthesizing questions helps us achieve this. We acknowledge that this setup represents a relatively favorable, or near-best-case, setting for the evaluated models, and that performance could be lower on questions for which the relevant evidence is absent from the reference materials.

We emphasize that our evaluation focuses on verifiability, which is distinct from clinical correctness. A fully supported claim is not necessarily clinically correct, complete, or safe. For example, the cited sources may be outdated, making claims based on them no longer fully accurate. Conversely, a claim that is not fully supported by its cited quote may still be clinically correct.

Besides, \texttt{deepseek-v4-flash} serves multiple roles in our automated evaluation pipeline, including as the LLM-based claim filter, claim support judge, and recovery probe, while also being evaluated as one of the answer-generation models. We incorporated alternative LLM judges (\texttt{gpt-5-mini}, \texttt{grok-4.3}) and did not observe evidence of self-preference (e.g., an LLM judge favoring answers generated by itself or by models from the same provider) by \texttt{deepseek-v4-flash}. However, we cannot fully disentangle whether the use of \texttt{deepseek-v4-flash} in the stages preceding the claim support judge stage affects its results as an evaluated model.

We also examined the sensitivity of our results to the generation prompt. By modifying the prompt to explicitly require quotes to support every point of a claim sentence, we observed improvements in claim support to varying degrees across most models, while citation coverage and verbatim compliance remained largely unchanged (Appendix~\ref{app:prompt-ablation}). Despite these improvements, the gap between attaching a verbatim quote and CCR remains, supporting our main conclusions in this paper.

Overall, our work provides a practical framework for evaluating the verifiability of clinical QA systems and highlights the gap between producing verbatim quotes and providing sufficient evidence to fully support clinical claims. These findings offer practical insights for developing verifiable clinical QA systems, selecting models for deployment, and provide a more fine-grained perspective on LLM benchmarking. We release our artifacts for future research to build on.

\clearpage
\makeatletter
\global\@colroom\@colht
\global\vsize\@colht
\makeatother

\iffinal
\acks{This work was supported by a Transdisciplinary Research and Collaboration (TRaC) Grant (No.~80070352) from the Welch Center for Prevention, Epidemiology, and Clinical Research at Johns Hopkins University.}
\fi

\bibliography{references}

\onecolumn
\appendix
% Appendix pages end where their content ends; no glue is stretched to fill them.
\raggedbottom

%%%%%%%%%%%%%%%%%%%%%%%%%%%%%%%%%%%%%%%%%%%%%%%%%%%%%%%%%%%%%%%%%%%%%%%%%%%%%%
%%  A.Additional Results and Implementation Details
%%%%%%%%%%%%%%%%%%%%%%%%%%%%%%%%%%%%%%%%%%%%%%%%%%%%%%%%%%%%%%%%%%%%%%%%%%%%%%
\section{Additional Results and Implementation Details}
\label{app:impl}

\subsection{Guideline Corpus Construction}
\label{app:corpus}

Each document is parsed along its native table of contents into one plain-text file per section, which yields the tree of Section~\ref{subsec:guideline}. A node records its identifier, section number and title, its parent and children, the plain-text file holding its prose, that file's length in characters, the figures and tables it contains, and a summary. Figures and tables are replaced by a transcription from \texttt{gpt-4o}, so every node is plain text. Each section then gets a summary of roughly 250 characters from
\texttt{gpt-4o-mini}, and the retrieval tool ranks over these summaries rather than the
section text. Both prompts are in Appendix~\ref{app:prompts-data}. Figure~\ref{fig:node} shows one node of the tree.

\begin{figure}[H]
\centering
\begin{lstlisting}[style=prompt]
...
{
  "section_id":     "sec-10-2",
  "number":         "6.2",
  "title":          "Hypertensive Emergencies and Severe Hypertension
                     in Nonpregnant and Nonstroke Patients",
  "parent_id":      "sec-10",
  "children":       ["sec-10-2-1"],
  "text_file":      "sections/sec-10-2.txt",
  "content_length": 5405,
  "figures":        [],
  "tables":         ["t56", "t26", "t27"],
  "summary":        "Covers management of hypertensive emergencies and
                     severe hypertension in nonpregnant, non-stroke
                     patients. Recommendations include ICU admission
                     for emergencies, gradual BP reduction strategies,
                     and cautious use of antihypertensives in
                     asymptomatic severe hypertension."
}
...
\end{lstlisting}
\caption{An example of a node of the guideline tree.}
\label{fig:node}
\end{figure}

\subsection{Synthetic Question Construction}
\label{app:question-set}

Questions are generated from the guideline tree of Appendix~\ref{app:corpus}, one guideline at a time. A per-guideline whitelist of clinical chapters excludes front matter, back matter and administrative chapters. The subsections of the retained chapters are then sampled at an even stride, capped at 25 per guideline, so that coverage spans the whole document. This yields 74 anchor subsections: 14 from the 2025 AHA guideline on high blood pressure, 25 from the 2019 AHA guideline on the primary prevention of cardiovascular disease, 25 from the 2026 ADA Standards of Care in Diabetes, and 10 from the 2026 AHA guideline on blood cholesterol.

Each anchor subsection then receives one question in each of three styles, giving $74 \times 3 = 222$ questions. \emph{Direct lookup} asks for a fact the subsection states, \emph{patient-specific scenario} poses a short patient case it answers, and \emph{shared decision-making} asks how its content would be discussed with a patient. The three prompts are given in Appendix~\ref{app:prompts-data}. The released dataset records each question's style, its source guideline, and the identifier of the subsection it came from. Figure~\ref{fig:question-styles} gives the three questions written from one subsection.

\begin{figure}[H]
\centering
\small
\begin{tcolorbox}[enhanced, colback=white, colframe=black!30,
  boxrule=0.4pt, left=6pt, right=6pt, top=5pt, bottom=5pt,
  title={\textsc{Direct lookup} \hfill \normalfont 2025 blood pressure guideline, Section 2.1},
  coltitle=black, colbacktitle=black!6, fonttitle=\bfseries\footnotesize]
How should blood pressure be categorized for adults to effectively prevent and treat high blood pressure according to the 2025 AHA guidelines?
\end{tcolorbox}

\smallskip
\begin{tcolorbox}[enhanced, colback=white, colframe=black!30,
  boxrule=0.4pt, left=6pt, right=6pt, top=5pt, bottom=5pt,
  title={\textsc{Patient-specific scenario} \hfill \normalfont 2025 blood pressure guideline, Section 2.1},
  coltitle=black, colbacktitle=black!6, fonttitle=\bfseries\footnotesize]
A 62-year-old man with a history of obesity and diabetes presents with a blood pressure of 145/90 mmHg. What classification should be applied to his blood pressure based on the current recommendations?
\end{tcolorbox}

\smallskip
\begin{tcolorbox}[enhanced, colback=white, colframe=black!30,
  boxrule=0.4pt, left=6pt, right=6pt, top=5pt, bottom=5pt,
  title={\textsc{Shared decision-making} \hfill \normalfont 2025 blood pressure guideline, Section 2.1},
  coltitle=black, colbacktitle=black!6, fonttitle=\bfseries\footnotesize]
How can I explain to my patient the importance of categorizing their blood pressure as normal, elevated, or hypertensive, and how this classification affects their treatment options?
\end{tcolorbox}

\caption{One question in each style, all three written from the same guideline subsection.}
\label{fig:question-styles}
\end{figure}

Here we report an earlier smaller-scale annotation result. In a previous version of the synthetic dataset with 135 questions generated using the same prompts, two clinical professionals independently reviewed all questions. One reviewer approved all 135 questions, while the other approved 130 and suggested wording edits for the remaining five.

\subsection{Retrieval Hit Rate}
\label{subsec:retrieval-results}
\label{subsec:hit-rate}

We compare our LLM-guided semantic retrieval against two baselines. The dense vector baseline embeds each section with OpenAI's \texttt{text-embedding-3-small} using a window of 24,000 characters. A section longer than one window is embedded window by window and the vectors mean-pooled. Questions are embedded the same way, and sections are ranked by cosine similarity. The lexical baseline is Okapi BM25 ($k_1 = 1.5$, $b = 0.75$) over the same section texts, lower-cased and tokenized into alphanumeric runs, with stopwords removed and Porter2 stemming applied. Each method is run over the question set of Section~\ref{subsec:dataset}.

We report the retrieval hit rate, the fraction of questions for which the ground-truth section is contained in the returned set. Let the evaluation set be $\{(q_i, s_i)\}_{i=1}^{N}$, where $N$ is the number of questions, $q_i$ is a question and $s_i$ is the section it was generated from, and let $R(q_i)$ be the set of sections a method returns for $q_i$. The hit rate is
\begin{equation}
  \mathrm{HR} = \frac{1}{N} \sum_{i=1}^{N} \mathbf{1}\!\left[\, s_i \in R(q_i) \,\right],
\end{equation}
where $\mathbf{1}[\cdot]$ is the indicator function. A hit therefore requires returning the exact section the question was written from.

The LLM-guided semantic retrieval is not deterministic and returns a variable number of sections, so we report each backing model over three independent runs, with the mean number of sections retrieved, the mean hit rate, and the range across runs. The lexical and dense baselines instead rank the whole section pool, so we report each of them at Top-3, Top-5 and Top-10. Table~\ref{tab:retrieval-hitrate} gives the full results.

\begin{table}[H]
\centering
\caption{Retrieval hit rate on the evaluation set. Every model runs the selector at its
provider's default temperature with thinking off, over three runs.}
\label{tab:retrieval-hitrate}
{\normalsize
\setlength{\tabcolsep}{4pt}
\begin{tabular}{@{}lcc@{}}
\toprule
Method & Avg.\ sections & Hit rate (\%) \\
\midrule
\csvreader[filter strcmp={\colgroup}{baseline}, late after line=\\]%
  {tables/hit_rate.csv}{group=\colgroup, model=\colmodel, sections=\colsec, hit=\colhit}%
  {\colmodel & \colsec & \colhit}
\midrule
\csvreader[%
  filter strcmp={\colmain}{1},
  late after line=\\]%
  {tables/hit_rate.csv}{vendor=\colvendor, model=\colmodel, sections=\colsec, hit=\colhit, lo=\collo, hi=\colhi, besthit=\colbesthit, maintable=\colmain}%
  {\vendormark{\colvendor}\texttt{\colmodel} & \colsec & \bestif{\colbesthit}{\colhit}\ {\tiny(\collo--\colhi)}}
\bottomrule
\end{tabular}}
\end{table}

Every backing model except \texttt{glm-4.7-flash} achieves a higher hit rate than either baseline at Top-5, while returning fewer than five sections. Within the same provider, larger models tend to perform better, yet lightweight models are already sufficient for the retrieval task. For instance, the performance gap between \texttt{deepseek-v4-flash} and \texttt{deepseek-v4-pro} is within one percentage point. This indicates that a brief description of each section's topic is enough to retrieve it, without the full text. We treat the section each synthesized question was drawn from as the sole ground-truth section, so a method that returns equally sufficient evidence from elsewhere is still scored as a miss. The reported hit rates are therefore conservative.

\subsection{System User Interface}
\label{app:system-ui}

\begin{figure}[H]
  \centering
  \includegraphics[width=\textwidth]{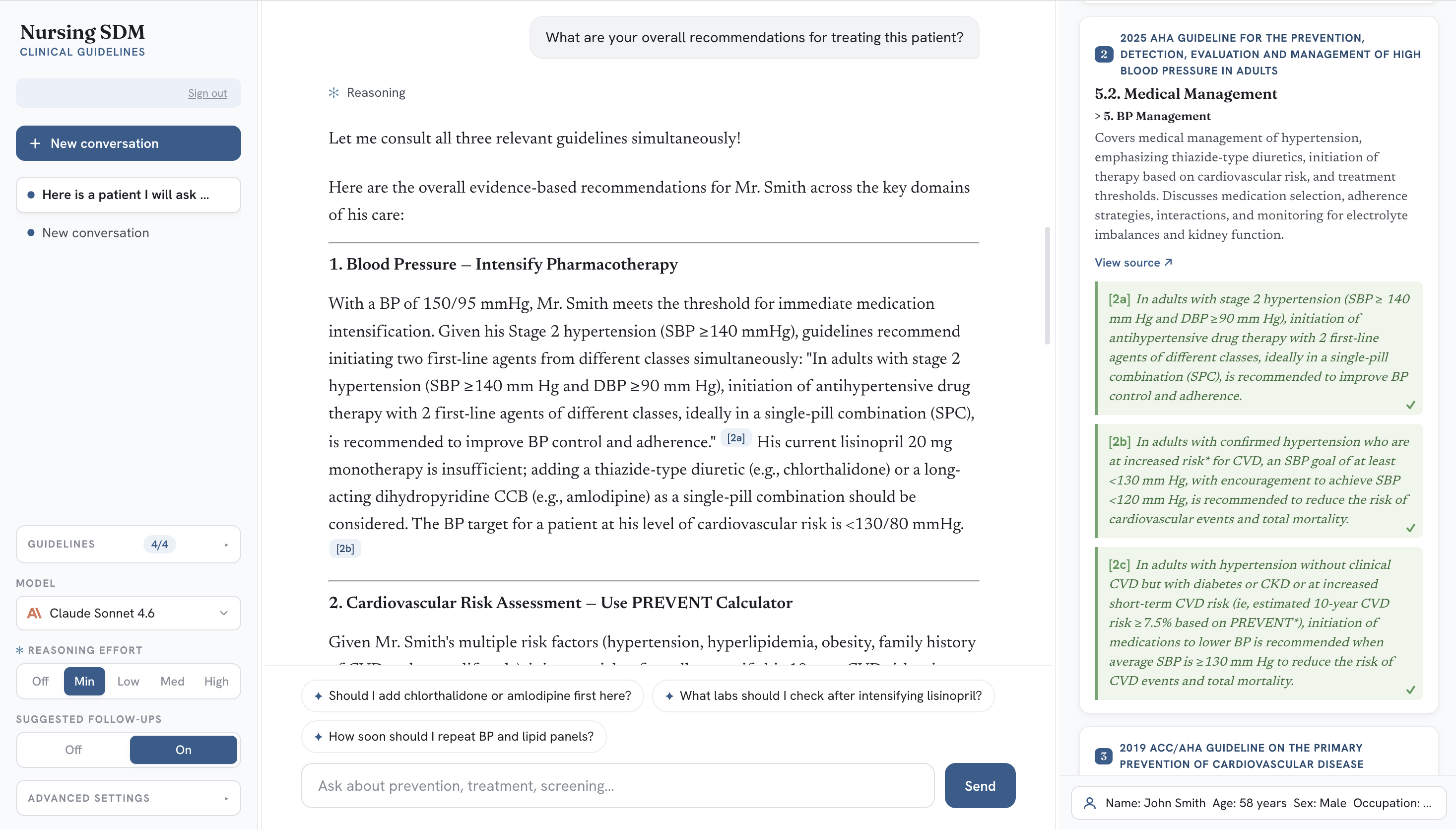}
  \caption{The system user interface.}
  \label{fig:system-ui}
\end{figure}

\subsection{Generation Model Configuration}
\label{app:gen-config}

Table~\ref{tab:gen-config} records what each generation model ran with. 

\begin{table}[H]
\centering
\caption{Generation configuration per model. }
\label{tab:gen-config}
{\small
\setlength{\tabcolsep}{4pt}
\begin{tabular}{@{}l|l@{}}
\toprule
\multicolumn{1}{@{}l}{Model} & \multicolumn{1}{l@{}}{Reasoning} \\
\midrule
\csvreader[late after line=\\]%
  {tables/gen_config.csv}{vendor=\colvendor, model=\colmodel, effort=\coleffort}%
  {\vendorcell{\colvendor}\texttt{\colmodel} & \coleffort}
\bottomrule
\end{tabular}}
\end{table}
By default, we choose high reasoning effort and the default temperature for all models, with three exceptions: \texttt{gpt-4.1-mini}, \texttt{claude-haiku-4.5}, and \texttt{glm-4.7-flash}.
Since \texttt{gpt-4.1-mini} is not a reasoning model, reasoning effort is not applicable for it.
\texttt{glm-4.7-flash} does not support the reasoning effort parameter and only allows reasoning to be enabled or disabled, and we chose to enable reasoning.
Similarly, \texttt{claude-haiku-4.5} does not support the reasoning effort parameter but supports extended thinking through the \texttt{thinking.budget\_tokens} parameter.
We used the value recommended by the \texttt{pydantic-ai} framework for high reasoning effort, which corresponds to 16,384 tokens.

\subsection{Evaluation Model Configuration}
\label{app:judge-config}

\begin{table}[H]
\centering
\caption{Configuration of the LLM stages of the evaluation pipeline.
\emph{default} means no temperature parameter was sent, so the provider's default value applied.}
\label{tab:judge-config}
{\small
\setlength{\tabcolsep}{4pt}
\begin{tabular}{@{}l|lll@{}}
\toprule
\multicolumn{1}{@{}l}{Component} & \multicolumn{1}{l}{Model}
  & \multicolumn{1}{l}{Temperature} & \multicolumn{1}{l@{}}{Reasoning} \\
\midrule
\csvreader[late after line=\\]%
  {tables/judge_config.csv}{component=\colcomponent, model=\colmodel,
   temp=\coltemp, thinking=\colthinking}%
  {\colcomponent & \texttt{\colmodel} & \coltemp & \colthinking}
\bottomrule
\end{tabular}}
\end{table}

\subsection{Verbatim Match Tiers}
\label{app:match-tiers}

Beyond an exact, character-for-character match, two relaxed tiers admit a quote. Figure~\ref{fig:tiers} shows one passing quote from each.

\begin{description}
  \item[Normalized.] The quote is a substring of the section after both are normalized.
Normalization puts Unicode in a canonical form, lower-cases, folds curly quotation marks and the dash variants onto their ASCII equivalents, collapses runs of whitespace to a single space, strips the guideline's own reference markers along with list bullets and table markup, and closes up the space those deletions leave before punctuation.
  \item[Elided.] The quote contains an ellipsis, and the fragments it separates each occur
in the section text in their original order. Fragments shorter than ten characters are disregarded and at least two must remain.
\end{description}

\begin{figure}[H]
\centering
\small
\begin{tcolorbox}[enhanced, colback=white, colframe=black!30,
  boxrule=0.4pt, left=6pt, right=6pt, top=5pt, bottom=5pt,
  title={\textsc{Normalized} \hfill \normalfont\texttt{claude-haiku-4.5}},
  coltitle=black, colbacktitle=black!6, fonttitle=\bfseries\footnotesize]
\textbf{Source.} \ub{F}asting is not required for Lp(a) testing\\
\textbf{Quote.} \ub{f}asting is not required for Lp(a) testing
\end{tcolorbox}

\smallskip
\begin{tcolorbox}[enhanced, colback=white, colframe=black!30,
  boxrule=0.4pt, left=6pt, right=6pt, top=5pt, bottom=5pt,
  title={\textsc{Elided} \hfill \normalfont\texttt{gpt-5.4-mini}},
  coltitle=black, colbacktitle=black!6, fonttitle=\bfseries\footnotesize]
\textbf{Source.} Weight loss of 5\% to 10\% of initial weight\ub{, achieved through
comprehensive lifestyle intervention,} has been shown to improve BP, delay the onset of
T2DM\\
\textbf{Quote.} Weight loss of 5\% to 10\% of initial weight\dots{} has been shown to
improve BP, delay the onset of T2DM
\end{tcolorbox}
\caption{Examples of the two relaxed tiers. Highlights mark what differs from the
source.}
\label{fig:tiers}
\end{figure}

\subsection{Claim Units}
\label{app:claim-units}

The smallest unit of analysis in this paper is the \emph{claim sentence}. The claim sentences are extracted by the following steps.

\begin{enumerate}
  \item \textbf{Flatten the markdown.} A heading or a ``1.'' bullet would otherwise reach the sentence splitter as prose and be counted as a sentence of its own.
  \item \textbf{Strip the citation markers}, recording where each one sat. They cannot survive the split, because the quote inside a marker carries its own punctuation and would create sentence boundaries that are not there; the recorded positions are what later attributes a citation to a sentence.
  \item \textbf{Split the remaining prose into sentences} with pySBD.
  \item \textbf{Drop anything shorter than 15 characters}, which is a fragment rather than
  a claim.
  \item \textbf{Apply the claim filter} (\texttt{deepseek-v4-flash}, prompt in
Appendix~\ref{app:prompts-claim-filter}), which decides which of the remaining sentences state a clinical claim. Those sentences are the claims.
\end{enumerate}

\subsection{The Attribution Window}
\label{app:window}

\begin{figure}[H]
\centering
{\small
\begin{minipage}[t]{0.44\textwidth}
\centering
\begin{tabular}{@{}l@{\hskip 10pt}ccccccc@{}}
\toprule
       & $s_1$ & $s_2$ & \mka & $s_3$ & $s_4$ & $s_5$ & \mkb \\
\midrule
$k=0$  & \cu & \ca & & \cu & \cu & \cb & \\
$k=1$  & \cu & \ca & & \cu & \cu & \cb & \\
$k=2$  & \ca & \ca & & \cu & \cb & \cb & \\
$k=3$  & \ca & \ca & & \cb & \cb & \cb & \\
\bottomrule
\end{tabular}

\vspace{5pt}
{\footnotesize (a) every sentence is a claim sentence}
\end{minipage}%
\hspace{0.06\textwidth}%
\begin{minipage}[t]{0.44\textwidth}
\centering
\begin{tabular}{@{}l@{\hskip 10pt}ccccccc@{}}
\toprule
       & $s_1$ & $s_2^{\dagger}$ & \mka & $s_3$ & $s_4^{\dagger}$ & $s_5$ & \mkb \\
\midrule
$k=0$  & \cu & \cn & & \cu & \cn & \cb & \\
$k=1$  & \ca & \cn & & \cu & \cn & \cb & \\
$k=2$  & \ca & \cn & & \cb & \cn & \cb & \\
$k=3$  & \ca & \cn & & \cb & \cn & \cb & \\
\bottomrule
\end{tabular}

\vspace{5pt}
{\footnotesize (b) $s_2$ and $s_4$ are not claim sentences}
\end{minipage}
}
\caption{The attribution window $k$: how a citation is credited to claim sentences. The answer runs $s_1, s_2, c_1, s_3, s_4, s_5, c_2$, each $s$ a sentence and each $c$ a marker labelled at the boundary it occupies, its arrow pointing to the side it credits. Fills: \swatch{terrA} credited by $c_1$, \swatch{terrB} by $c_2$, \swatch{uncov} a claim sentence no marker credits, \swatch{noncl} ($\dagger$) a sentence that makes no claim. A marker credits only sentences at or before its own position, so $c_1$ never credits beyond $s_2$ however large $k$ grows. It also never reaches back past the previous marker, which stops $c_2$ at $s_3$ in (a). Because $k$ counts claim sentences, in (b) $c_2$ passes over $s_4$ to reach $s_3$.}
\label{fig:window}
\end{figure}

\subsection{Full Citation Coverage Results}
\label{app:coverage-full}

\begin{table}[H]
\centering
\caption{Citation coverage by generation model, over the full ladder of attribution windows. Each figure counts every citation the model emitted, with the verbatim-passing subset in parentheses. \emph{Claims} is the denominator of every rate.}
\label{tab:coverage-full}
{\footnotesize
\setlength{\tabcolsep}{3pt}
\begin{tabular}{@{}lc|cccccc@{}}
\toprule
\multicolumn{1}{@{}l}{} & \multicolumn{1}{c}{}
  & \multicolumn{6}{c}{Coverage (\%), attribution window} \\
\cmidrule(lr){3-8}
\multicolumn{1}{@{}l}{Model} & \multicolumn{1}{c}{Claims}
  & \multicolumn{1}{c}{0} & 1 & 2 & 3 & 4 & 5 \\
\midrule
\csvreader[late after line=\\]%
  {tables/coverage.csv}{model=\colmodel, claims=\colclaims,
   cc0=\cczero, cc1=\ccone, cc2=\cctwo, cc3=\ccthree, cc4=\ccfour, cc5=\ccfive,
   cc0c=\cczeroc, cc1c=\cconec, cc2c=\cctwoc, cc3c=\ccthreec, cc4c=\ccfourc, cc5c=\ccfivec}%
  {\texttt{\colmodel} & \colclaims
   & \cczero\,\gatedsm{\cczeroc}
   & \ccone\,\gatedsm{\cconec}
   & \cctwo\,\gatedsm{\cctwoc}
   & \ccthree\,\gatedsm{\ccthreec}
   & \ccfour\,\gatedsm{\ccfourc}
   & \ccfive\,\gatedsm{\ccfivec}}
\bottomrule
\end{tabular}}
\end{table}

\subsection{Citation Coverage Failure Cases}
\label{app:coverage-failures}

\begin{figure}[H]
\centering
\small
\begin{tcolorbox}[enhanced, colback=white, colframe=black!30,
  boxrule=0.4pt, left=6pt, right=6pt, top=5pt, bottom=5pt,
  title={\textsc{No citation at all} \hfill \normalfont\texttt{glm-4.7-flash}},
  coltitle=black, colbacktitle=black!6, fonttitle=\bfseries\footnotesize]
\textbf{Answer.} \ub{For adults with hypertension and chronic kidney disease, the guideline recommends a BP target of less than 130/80 mm Hg for treatment.}
\end{tcolorbox}

\smallskip
\begin{tcolorbox}[enhanced, colback=white, colframe=black!30,
  boxrule=0.4pt, left=6pt, right=6pt, top=5pt, bottom=5pt,
  title={\textsc{An uncited tail} \hfill \normalfont\texttt{claude-haiku-4.5}},
  coltitle=black, colbacktitle=black!6, fonttitle=\bfseries\footnotesize]
\textbf{Answer.} For a 50-year-old man with confirmed hypertension and chronic kidney disease, the target blood pressure should be less than 130/80 mm Hg. [1] \ub{This is a Class I recommendation with strong supporting evidence for systolic BP and expert opinion on diastolic BP.}
\end{tcolorbox}

\smallskip
\begin{tcolorbox}[enhanced, colback=white, colframe=black!30,
  boxrule=0.4pt, left=6pt, right=6pt, top=5pt, bottom=5pt,
  title={\textsc{Every citation at the end} \hfill \normalfont\texttt{gpt-4.1-mini}},
  coltitle=black, colbacktitle=black!6, fonttitle=\bfseries\footnotesize]
\textbf{Answer.} \ub{For a 45-year-old woman with obesity (BMI 32), type 2 diabetes, and low HDL-C, lifestyle management including a healthy diet, regular physical activity, and weight reduction of 5\% to 10\% is foundational to improve dyslipidemia and reduce cardiovascular risk.} \ub{Moderate-intensity statin therapy is indicated to achieve at least a 30\% to 49\% reduction in LDL-C and target \mbox{LDL-C \textless 100\,mg/dL} and \mbox{non--HDL-C \textless 130\,mg/dL}.} \ub{\dots{}} Referral to a registered dietitian nutritionist is also beneficial for dietary counseling in diabetes and dyslipidemia management. [1] [2] [3] [4] [5]
\end{tcolorbox}

\caption{One answer per coverage failure mode. Yellow highlights indicate sentences that are not covered by any citation ($k = 1$).}
\label{fig:coverage-failures}
\end{figure}

\subsection{Claim and Quote Statistics}
\label{app:support-length}
\begin{table}[H]
\centering
\caption{Claim support inside each claim-length band, in words of the claim sentence. \emph{Str.}
is the strict rate and \emph{Len.} the lenient rate of Table~\ref{tab:claimsupport}, computed over
the pairs in that band; $n$ is those pairs. Judged by \texttt{deepseek-v4-flash}.}
\label{tab:support-length}
{\small
\setlength{\tabcolsep}{3pt}
\begin{tabular}{@{}l|ccc|ccc|ccc|cc@{}}
\toprule
\multicolumn{1}{@{}l}{}
  & \multicolumn{3}{c}{$<25$ words}
  & \multicolumn{3}{c}{25--34 words}
  & \multicolumn{3}{c}{$\geq35$ words}
  & \multicolumn{2}{c}{All claims} \\
\cmidrule(lr){2-4} \cmidrule(lr){5-7} \cmidrule(lr){8-10} \cmidrule(l){11-12}
\multicolumn{1}{@{}l}{Model}
  & $n$ & Str. & \multicolumn{1}{c}{Len.}
  & $n$ & Str. & \multicolumn{1}{c}{Len.}
  & $n$ & Str. & \multicolumn{1}{c}{Len.}
  & Str. & \multicolumn{1}{c}{Len.} \\
\midrule
\csvreader[late after line=\\]%
  {tables/support_by_length.csv}{vendor=\colvendor, model=\colmodel,
   n1=\cola, s1=\colb, l1=\colc, n2=\cold, s2=\cole, l2=\colf,
   n3=\colg, s3=\colh, l3=\coli, strict=\colstrict, lenient=\collenient}%
  {\vendorcell{\colvendor}\texttt{\colmodel}
   & \cola & \colb & \colc
   & \cold & \cole & \colf
   & \colg & \colh & \coli
   & \colstrict & \collenient}
\bottomrule
\end{tabular}}
\end{table}

\begin{table}[H]
\centering
\caption{How much each model claims and how much it quotes. Judged by \texttt{deepseek-v4-flash}.}
\label{tab:claim-quote-stats}
{\small
\setlength{\tabcolsep}{3pt}
\begin{tabular}{@{}l|cccc|cccc|c|cc@{}}
\toprule
\multicolumn{1}{@{}l}{}
  & \multicolumn{4}{c}{Count ($n$)}
  & \multicolumn{4}{c}{Mean length (characters)}
  & \multicolumn{1}{c}{}
  & \multicolumn{2}{c}{Rate (\%)} \\
\cmidrule(lr){2-5} \cmidrule(lr){6-9} \cmidrule(l){11-12}
\multicolumn{1}{@{}l}{Model}
  & Claims & Claims$'$ & Quotes & \multicolumn{1}{c}{Quotes$'$}
  & Claims & Claims$'$ & Quotes & \multicolumn{1}{c}{Quotes$'$}
  & \multicolumn{1}{c}{Q$'$/C$'$}
  & Strict & \multicolumn{1}{c}{CCR} \\
\midrule
\csvreader[late after line=\\]%
  {tables/claim_quote_stats.csv}{vendor=\colvendor, model=\colmodel,
   claims=\colclaims, claimsp=\colclaimsp, quotes=\colquotes, quotesp=\colquotesp,
   claimchars=\colclaimch, claimpchars=\colclaimpch,
   quotechars=\colquotech, quotepchars=\colquotepch,
   density=\coldensity, strict=\colstrict, ccr=\colccr}%
  {\vendorcell{\colvendor}\texttt{\colmodel}
   & \colclaims & \colclaimsp & \colquotes & \colquotesp
   & \colclaimch & \colclaimpch & \colquotech & \colquotepch
   & \coldensity
   & \colstrict & \colccr}
\bottomrule
\end{tabular}}

\vspace{4pt}
\begin{minipage}{\textwidth}
\small
A prime marks the verbatim-compliant subset: \emph{Claims}$'$ are the claims carrying at least one quote that passes the verbatim check, and \emph{Quotes}$'$ are those quotes, so the primed pair is the population the support judge reads. \emph{Claims} is the denominator of the certified claim rate and \emph{Quotes} counts every citation marker the model emitted. Lengths are characters of the text as written. \emph{Q$'$/C$'$} is the mean number of compliant quotes (\emph{Quotes}$'$) one compliant claim (\emph{Claims}$'$) cites.
\end{minipage}
\end{table}

\subsection{Run-to-Run Variance}
\label{app:round-variance}

Every figure in the main text comes from one sampling round. To assess run-to-run variability, we ran all evaluated models twice more under the identical configuration (i.e., each model generated answers to the same questions in two additional rounds, followed by the same evaluation pipeline). The analysis in the main text is based on the results from round 1, and the additional rounds continue to support our main conclusions.

\begin{table}[H]
\centering
\caption{The same configuration, three times, as round 1 / round 2 / round 3. The columns are those of Table~\ref{tab:claim-funnel}.}
\label{tab:round-variance}
\scalebox{0.88}{%
{\footnotesize
\setlength{\tabcolsep}{3pt}
\begin{tabular}{@{}l|cc|cccc|c@{}}
\toprule
\multicolumn{1}{@{}l}{Model} & Sent. & \multicolumn{1}{c}{Claims}
  & Covered & Verbatim & Lenient & \multicolumn{1}{c}{CCR}
  & \multicolumn{1}{c}{Avg. CCR} \\
\midrule
\csvreader[late after line=\\]%
  {tables/round_variance.csv}{vendor=\colvendor, model=\colmodel,
   sent1=\sa, sent2=\sb, sent3=\sc, claims1=\ca, claims2=\cb, claims3=\cc,
   cov1=\va, cov2=\vb, cov3=\vc, verb1=\ba, verb2=\bb, verb3=\bc,
   len1=\la, len2=\lb, len3=\lc, ccr1=\ra, ccr2=\rb, ccr3=\rc, ccrmean=\ravg}%
  {\vendorcell{\colvendor}\texttt{\colmodel}
   & \sa / \sb / \sc & \ca / \cb / \cc
   & \va / \vb / \vc & \ba / \bb / \bc
   & \la / \lb / \lc & \ra / \rb / \rc
   & \textbf{\ravg}}
\bottomrule
\end{tabular}}%
}
\end{table}

\subsection{Claim Support Under Additional Judges}
\label{app:claim-support-second}

\begin{table}[H]
\centering
\caption{Table~\ref{tab:claimsupport} under the second judge, \texttt{gpt-5-mini}. $\kappa$ is Cohen's $\kappa$ between this judge and \texttt{deepseek-v4-flash} over three classes: \emph{fully supported}, \emph{partially supported}, and \emph{not supported}/\emph{contradicted}.}
\label{tab:claimsupport-second-gated}
{\small
\setlength{\tabcolsep}{3pt}
\begin{tabular}{@{}l|c|cccc|cccc|cc|c@{}}
\toprule
\multicolumn{1}{@{}l}{} & \multicolumn{1}{c}{}
  & \multicolumn{4}{c}{Shortfall ($n$)}
  & \multicolumn{4}{c}{Verdict ($n$)}
  & \multicolumn{2}{c}{Rate (\%)}
  & \multicolumn{1}{c}{Agree.} \\
\cmidrule(lr){3-6} \cmidrule(lr){7-10} \cmidrule(lr){11-12} \cmidrule(l){13-13}
\multicolumn{1}{@{}l}{Model} & \multicolumn{1}{c}{Cited}
  & Add. & Causal & Scope & \multicolumn{1}{c}{Strength}
  & Full & Partial & None & \multicolumn{1}{c}{Contra}
  & Strict & \multicolumn{1}{c}{Lenient}
  & $\kappa$ \\
\midrule
\csvreader[late after line=\\, filter strcmp={\coljudge}{gpt-5-mini}]%
  {tables/claim_support_gated.csv}{judge=\coljudge, vendor=\colvendor, model=\colmodel,
   paired=\colpaired, beststrict=\bstrict, bestlenient=\blenient,
   strict=\colstrict, lenient=\collenient,
   nfully=\colfull, npartial=\colpart, nnot=\colnot, ncontra=\colcontra,
   naddition=\coladd, ncausal=\colcausal, nscope=\colscope, nstrength=\colstrength,
   kappa3=\colkappa}%
  {\vendorcell{\colvendor}\texttt{\colmodel} & \colpaired
   & \coladd & \colcausal & \colscope & \colstrength
   & \colfull & \colpart & \colnot & \colcontra
   & \bestif{\bstrict}{\colstrict} & \bestif{\blenient}{\collenient}
   & \colkappa}
\bottomrule
\end{tabular}}
\end{table}

\begin{table}[H]
\centering
\caption{Table~\ref{tab:claimsupport} under the third judge, \texttt{grok-4.3}. $\kappa$ is Cohen's $\kappa$ between this judge and \texttt{deepseek-v4-flash} over three classes: \emph{fully supported}, \emph{partially supported}, and \emph{not supported}/\emph{contradicted}.}
\label{tab:claimsupport-third-gated}
{\small
\setlength{\tabcolsep}{3pt}
\begin{tabular}{@{}l|c|cccc|cccc|cc|c@{}}
\toprule
\multicolumn{1}{@{}l}{} & \multicolumn{1}{c}{}
  & \multicolumn{4}{c}{Shortfall ($n$)}
  & \multicolumn{4}{c}{Verdict ($n$)}
  & \multicolumn{2}{c}{Rate (\%)}
  & \multicolumn{1}{c}{Agree.} \\
\cmidrule(lr){3-6} \cmidrule(lr){7-10} \cmidrule(lr){11-12} \cmidrule(l){13-13}
\multicolumn{1}{@{}l}{Model} & \multicolumn{1}{c}{Cited}
  & Add. & Causal & Scope & \multicolumn{1}{c}{Strength}
  & Full & Partial & None & \multicolumn{1}{c}{Contra}
  & Strict & \multicolumn{1}{c}{Lenient}
  & $\kappa$ \\
\midrule
\csvreader[late after line=\\, filter strcmp={\coljudge}{grok-4.3}]%
  {tables/claim_support_gated.csv}{judge=\coljudge, vendor=\colvendor, model=\colmodel,
   paired=\colpaired, beststrict=\bstrict, bestlenient=\blenient,
   strict=\colstrict, lenient=\collenient,
   nfully=\colfull, npartial=\colpart, nnot=\colnot, ncontra=\colcontra,
   naddition=\coladd, ncausal=\colcausal, nscope=\colscope, nstrength=\colstrength,
   kappa3=\colkappa}%
  {\vendorcell{\colvendor}\texttt{\colmodel} & \colpaired
   & \coladd & \colcausal & \colscope & \colstrength
   & \colfull & \colpart & \colnot & \colcontra
   & \bestif{\bstrict}{\colstrict} & \bestif{\blenient}{\collenient}
   & \colkappa}
\bottomrule
\end{tabular}}
\end{table}

\subsection{Human Annotation Against the Judge}
\label{app:human-annotation}

Two annotators labelled a stratified sample of 70 claim/quote pairs by hand: 30 the LLM judge (\texttt{deepseek-v4-flash}) called fully supported, 30 partially supported and 10 neither. Each pair below reports an unweighted Cohen's $\kappa$ and a quadratic-weighted one, both over three classes: \emph{fully supported}, \emph{partially supported}, and \emph{not supported}/\emph{contradicted}. The three classes are ordered, so confusing the two ends is a worse error than confusing neighbours.

\newcommand{\humanmatrix}[1]{%
\csvreader[filter strcmp={\colpair}{#1}]%
  {tables/human_agreement.csv}%
  {pair=\colpair, first=\hmfirst, second=\hmsecond, n=\hmn,
   ff=\hmff, fp=\hmfp, fn=\hmfn, pf=\hmpf, pp=\hmpp, pn=\hmpn,
   nf=\hmnf, np=\hmnp, nn=\hmnn,
   raw=\hmraw, kappa=\hmkappa, wkappa=\hmwkappa,
   lenient=\hmlen, stricter=\hmstr, far=\hmfar}%
  {\begin{tabular}{@{}l|ccc@{}}
   \multicolumn{4}{@{}l}{\hmfirst\ against \hmsecond} \\[2pt]
   \toprule
    & Fully & Part. & Neither \\
   \midrule
   Fully   & \hmff & \hmfp & \hmfn \\
   Part.   & \hmpf & \hmpp & \hmpn \\
   Neither & \hmnf & \hmnp & \hmnn \\
   \bottomrule
   \multicolumn{4}{@{}l}{$\kappa$ \hmkappa, weighted \hmwkappa} \\
   \multicolumn{4}{@{}l}{agreement \hmraw\% of \hmn} \\
   \end{tabular}}%
}

\begin{table}[H]
\centering
\caption{The three rater pairs on the same 70 pairs. Neither stands for \emph{not supported} or \emph{contradicted}.}
\label{tab:human-agreement}
{\small
\humanmatrix{a1-judge}\hfill\humanmatrix{a2-judge}\hfill\humanmatrix{a1-a2}
}
\end{table}

\subsection{Recovery from the Read Set}
\label{app:section-lookup-main}

Recovery runs in two steps. An LLM judge (\texttt{deepseek-v4-flash}) is shown the question, the claim and every guideline section the model read, and asked to copy out a span that states the part of the claim the quotes leave unsupported. That span is then re-checked against the source with the verbatim matcher of Section~\ref{subsec:vcr}, and only spans that pass the check are counted.

\begin{table}[H]
\centering
\caption{Recovery by evaluated model. \emph{Cases} is the claims the support judge marked
partially supported with an unsupported addition; \emph{Recov.} is how many of them the
probe returned a span for that passed the verbatim re-check, and \emph{Rate} is that share.
Judged by \texttt{deepseek-v4-flash}.}
\label{tab:sectionlookup}
{\small
\setlength{\tabcolsep}{4pt}
\begin{tabular}{@{}lcc|c@{}}
\toprule
\multicolumn{1}{@{}l}{Model} & \multicolumn{1}{c}{Cases}
  & \multicolumn{1}{c}{Recov.} & \multicolumn{1}{c}{Rate (\%)} \\
\midrule
\csvreader[late after line=\\, filter strcmp={\colmain}{1}]%
  {tables/section_lookup.csv}{maintable=\colmain, vendor=\colvendor, model=\colmodel,
   probed=\colcases, recoveredver=\colrecov, recoveredverpct=\colloc}%
  {\vendorcell{\colvendor}\texttt{\colmodel} & \colcases & \colrecov & \colloc}
\bottomrule
\end{tabular}}
\end{table}

\subsection{Generation Prompt Sensitivity}
\label{app:prompt-ablation}

Every figure in the main text comes from the single generation prompt of Appendix~\ref{app:prompts-generation}. To observe how much of our results depends on that prompt, we re-ran all evaluated models with one instruction changed and nothing else: the description of what a quote should be. The default asks for ``a SHORT verbatim passage copied from the section text''; the variant asks for ``a verbatim passage copied from the section text that supports EVERY point the claim makes''. 

Strict claim support rises for every model under the variant, by 11.3 percentage points on average, while citation coverage and verbatim compliance remain almost unchanged.
We also observed that the average quote length increased for all 14 models under the new prompt, with a pooled increase of $1.27\times$ relative to the original prompt.
These findings suggest that explicitly instructing models to provide sufficient supporting evidence can improve CCR (and strict claim support rates) to some extent, while our main conclusions in this paper remain unchanged.

\begin{table}[H]
\centering
\caption{The generation prompt varied in one instruction. The columns are those of Table~\ref{tab:claim-funnel}, and \emph{$\Delta$CCR} is the change against the same model's round~1.}
\label{tab:prompt-ablation}
\scalebox{0.88}{%
{\footnotesize
\setlength{\tabcolsep}{3pt}
\begin{tabular}{@{}l|cc|cccc|c@{}}
\toprule
\multicolumn{1}{@{}l}{Model} & Sent. & \multicolumn{1}{c}{Claims}
  & Covered & Verbatim & Lenient & \multicolumn{1}{c}{CCR}
  & \multicolumn{1}{c}{$\Delta$CCR} \\
\midrule
\csvreader[late after line=\\]%
  {tables/prompt_ablation.csv}{vendor=\colvendor, model=\colmodel,
   sent=\colsent, claims=\colclaims, cov=\colcov, verb=\colverb,
   len=\collen, ccr=\colccr, dccr=\coldccr}%
  {\vendorcell{\colvendor}\texttt{\colmodel}
   & \colsent & \colclaims & \colcov & \colverb & \collen & \colccr
   & \textbf{\coldccr}}
\bottomrule
\end{tabular}}%
}
\end{table}

%%%%%%%%%%%%%%%%%%%%%%%%%%%%%%%%%%%%%%%%%%%%%%%%%%%%%%%%%%%%%%%%%%%%%%%%%%%%%%
%%  B.Prompts
%%%%%%%%%%%%%%%%%%%%%%%%%%%%%%%%%%%%%%%%%%%%%%%%%%%%%%%%%%%%%%%%%%%%%%%%%%%%%%
\clearpage
\section{Prompts}
\label{app:prompts}

\subsection{Corpus and Dataset Construction}
\label{app:prompts-data}

The first two prompts build the corpus of Section~\ref{subsec:guideline}; the rest
synthesize the question set of Section~\ref{subsec:dataset}, one prompt per question
style.

\begin{promptbox}{Figure and table transcription}
\begin{lstlisting}[style=prompt]
# Visual-to-Text Transcription Prompt

You will receive a visual (image) from a clinical guideline together with its official caption. **Produce a structured text transcription of the image content** so a reader who never sees the original could reconstruct it.

## Mission

Create a description so complete that someone could **reconstruct the equivalent image** without seeing the original.

**Document, don't interpret.** Record objectively—do not summarize, infer, or fill in clinical reasoning the visual does not explicitly show.

You will be given the caption separately by the assembler; **do not repeat the caption text** in your output. Focus on the visual content.

---

## Core Principles

1. **Completeness** — capture every word, number, symbol, visual element, and spatial relationship
2. **Objectivity** — describe connections and elements without explaining meaning
3. **Structure preservation** — maintain original organization (hierarchy, sequence, network, spatial)
4. **Reconstructability test** — could someone redraw this from your description?

---

## Description Framework

### 1. Document structure
State the fundamental organization on the first line:
```
Structure: [table / flowchart / decision-tree / diagram / grid / nomogram / etc.]
```

### 2. Define navigation
- **Grid-like**: rows/columns or labeled sections
- **Flowcharts**: flow direction (top→bottom, left→right) and entry point
- **Spatial**: regions (top-left, center, bottom-right, etc.)

### 3. Enumerate elements
```
[Element Type] [Position]: [Exact Content]
  - Visual: [shape, color, style if meaningful]
  - Contains: [sub-elements if applicable]
```

### 4. Document connections
- **Arrows**: direction, style, endpoints
- **Lines**: type, endpoints, style
- **Containment**: nested relationships
- **Alignment**: shared rows/columns
- **Grouping**: visual clusters

### 5. Preserve visual semantics
Document any coding systems used:
- Color coding (e.g., green = Class I recommendation)
- Shape coding (e.g., diamond = decision node)
- Line style (e.g., dashed = alternative path)

### 6. Include annotations
Footnotes, legends, axis labels, units, citations, margins/headers.

---

## Medical Precision Requirements

1. **Numerical values and units** — "≥190 mg/dL (≥4.9 mmol/L)", not "high level"
2. **Comparison operators** — preserve: ≥, >, <, ≤, =, ≠
3. **Drug names** — exact spelling and capitalization
4. **Abbreviations** — keep verbatim: "LDL-C", "ASCVD", "PCSK9"
5. **Recommendation language** — exact: "is recommended", "is reasonable"
6. **Class/Level designations** — "(Class I)", "(Level A)" verbatim
7. **Temporal relationships** — "before", "after", "during"
8. **Range expressions** — distinguish: "10-20" vs "10–20" vs "10—20"
9. **Special symbols** — accurate: →, ⇄, ±, ×, ÷, ≈, ∆

---

## Output Format

Plain Markdown. Use headings (`##`, `###`) and bullet lists where useful for structure, but do not wrap the whole transcription in a code fence.

Begin directly with the `Structure:` line. Do not preface with "Here is a description..." or similar conversational filler.

---

## Quality Checklist (verify before finishing)

- [ ] Every visible element documented
- [ ] All text transcribed exactly (including footnotes and abbreviation legends)
- [ ] Navigation system clear
- [ ] All connections described
- [ ] Visual semantics explained where present
- [ ] Reconstructable from description alone
- [ ] No interpretation added
- [ ] Medical terms exact

---

## Prohibitions

- DO NOT repeat the caption (already stored separately)
- DO NOT summarize or paraphrase content
- DO NOT interpret clinical meaning
- DO NOT modify terminology
- DO NOT assume visual semantics the visual does not declare
- DO NOT skip structural relationships

---

You are a precise transcriber creating a blueprint for visual reconstruction, RAG retrieval, and clinical reference.

**Objectivity + Completeness = Success.**
\end{lstlisting}
\end{promptbox}

\begin{promptbox}{Section summarization}
\begin{lstlisting}[style=prompt]
You are creating a summary for a clinical guideline section. This summary will be used by a retrieval system to decide whether the section is relevant to a user's clinical question.

Requirements:
- STRICT LIMIT: max 250 characters (count carefully!)
- Create a COMPLETE, self-contained summary that captures the overall scope of this section.
- Focus on WHAT this section covers: key topics, patient populations, recommendations, and clinical scenarios addressed.
- Use clinical terminology appropriate for healthcare professionals.
- The summary must be COMPLETE — avoid being cut off mid-sentence.

Example:
"Addresses statin adverse effects including muscle symptoms, liver elevation, and diabetes risk with management strategies. Provides guidance on patient communication and rechallenge protocols."

Output the summary text only — no preface, no quotes, no labels.
\end{lstlisting}
\end{promptbox}

\begin{promptbox}{Question generation: direct lookup}
\begin{lstlisting}[style=prompt]
You are simulating realistic queries that nurses and multidisciplinary care team members would type into an AI clinical support agent during patient care. Generate direct, fact-seeking questions a care team member would ask to quickly retrieve guideline recommendations.

GUIDELINE: {guideline_name}

SECTION (breadcrumb, root → current node):
{title_path}

CONTENT (this section's text, with figure/table descriptions resolved inline):
{content}

EXAMPLES of the question style:
- According to current 2018 guidelines on the management of blood cholesterol, what are the recommended management options for patients 40 to 75 years of age with diabetes mellitus and LDL-C ≥70 mg/dL (≥1.8 mmol/L)?
- What lifestyle modifications and medication options are recommended by hypertension guidelines for adults with an average blood pressure ≥140/90 mm Hg, and for selected adults with an average blood pressure ≥130/80 mm Hg who have clinical cardiovascular disease, prior stroke, diabetes, chronic kidney disease, or a 10-year predicted cardiovascular risk ≥7.5%?

Generate {num_questions} questions that:
1. Are grounded entirely in the provided CONTENT — do not require external knowledge
2. Have a definitive answer traceable to the guideline text in this specific section
3. Are **uniquely locatable to THIS section** — the question should carry enough specific detail (the particular population, threshold, drug, or scenario this section covers) that it could NOT be answered just as well from a sibling section or from the parent chapter's generic overview. This is what makes retrieval testable.
4. Cover a range of different topics and recommendations within this section — avoid asking about the same concept twice
5. Do NOT copy this section's heading verbatim — phrase the question in your own words (using the core disease or clinical term itself is fine; it's the verbatim heading to avoid)

Return ONLY a valid JSON array with this structure:
[
  {{"question": "Question text here"}}
]

Do NOT include any markdown formatting or code blocks, just the raw JSON array.
\end{lstlisting}
\end{promptbox}

\begin{promptbox}{Question generation: patient scenario}
\begin{lstlisting}[style=prompt]
You are simulating realistic queries that nurses and multidisciplinary care team members would type into an AI clinical support agent when consulting it about a specific patient. Generate clinical scenario questions that describe a single concrete patient and ask what the guideline recommends for them.

GUIDELINE: {guideline_name}

SECTION (breadcrumb, root → current node):
{title_path}

CONTENT (this section's text, with figure/table descriptions resolved inline):
{content}

What makes a good scenario question here:
- **One concrete patient, singular.** Describe an individual (age, sex, and the clinically relevant details), not a population or a generic class. No "patients who…"; instead "a 58-year-old man who…".
- **Make the patient fit THIS section's scope.** Every section applies to some specific situation — it may be a lab value or vital crossing a threshold, an age group, a particular comorbidity or prior event, a pregnancy or peri-procedural state, a treatment already underway, etc. Choose the patient's details so that they land squarely within (or right at the boundary of) the exact population, category, or condition this section governs. Pick whichever discriminating criteria this section actually uses — don't force a number if the section isn't numeric.
- **Application, not recitation.** The question should require applying the section's recommendation to this patient — classifying them, choosing the option, setting the target, deciding the next step — so it can't be answered by quoting a definition verbatim.

Generate {num_questions} scenario questions that:
1. Are answerable entirely from the provided CONTENT — no outside knowledge needed
2. Are settled by THIS section, not a sibling section or the parent chapter's overview — the patient's specific details are what point to this section
3. Read like something a nurse or care team member would actually ask with this patient in front of them
4. Vary the patient profile and clinical situation across questions — avoid structurally similar scenarios
5. Do NOT copy this section's heading verbatim — describe the clinical situation in your own words (using the core disease or clinical term itself is fine; it's the verbatim heading to avoid)

Return ONLY a valid JSON array with this structure:
[
  {{"question": "Question text here"}}
]

Do NOT include any markdown formatting or code blocks, just the raw JSON array.
\end{lstlisting}
\end{promptbox}

\begin{promptbox}{Question generation: shared decision-making}
\begin{lstlisting}[style=prompt]
You are simulating realistic queries that nurses and multidisciplinary care team members would type into an AI clinical support agent when preparing for or conducting a Shared Decision-Making (SDM) conversation with a patient. Generate questions focused on how to communicate options, counsel patients, or support the SDM process.

GUIDELINE: {guideline_name}

SECTION (breadcrumb, root → current node):
{title_path}

CONTENT (this section's text, with figure/table descriptions resolved inline):
{content}

EXAMPLES of the question style:
- My patient is a 62-year-old female with diabetes and high LDL who has struggled to maintain lifestyle changes. How should I approach the conversation about starting statin therapy?
- A 58-year-old male patient with obesity and uncontrolled hypertension is resistant to making dietary changes. What key points should I cover to help him understand the connection between his weight and blood pressure?
- During a follow-up visit, a patient asks why she needs to take medication if her blood pressure is only slightly elevated. What should I explain to help her understand the benefits and trade-offs of starting treatment?

Generate {num_questions} SDM-oriented questions that:
1. Focus on the care team's role in discussing options with patients — e.g., what to explain, how to approach a conversation, how to address patient concerns
2. Are directly grounded in the provided CONTENT
3. Are **uniquely locatable to THIS section** — the communication challenge should hinge on the specific options, trade-offs, or counseling points this section covers, so the answer is in this section and not a sibling section or the parent chapter's overview. This is what makes retrieval testable.
4. Cover a range of different SDM situations from this section — vary the communication challenge (e.g., initiating treatment, addressing reluctance, explaining trade-offs, supporting adherence)
5. Do NOT copy this section's heading verbatim — phrase the question in your own words (using the core disease or clinical term itself is fine; it's the verbatim heading to avoid)

Return ONLY a valid JSON array with this structure:
[
  {{"question": "Question text here"}}
]

Do NOT include any markdown formatting or code blocks, just the raw JSON array.
\end{lstlisting}
\end{promptbox}

\subsection{Retrieval}
\label{app:prompts-retrieval}

The LLM-guided semantic retrieval prompt of Section~\ref{subsec:semantic-retrieval}.

\begin{promptbox}{Section selection}
\begin{lstlisting}[style=prompt]
You are a clinical-guidelines retrieval selector. Given a question and a list of guideline sections (each: [doc_id] breadcrumb — summary), pick the sections whose text most likely contains what's needed to answer.

- Select 1-3 sections for a focused question on a single topic.
- Select up to 6 sections when the question spans multiple guidelines or clinical topics.
- Prefer sections that directly address the question, but also include closely supporting sections — the answer is often spread across sibling sections.
- Use ONLY doc_ids from the list; copy them verbatim. Order most-relevant first.
- Briefly say why in `reasoning`.
\end{lstlisting}
\end{promptbox}

\subsection{Generation}
\label{app:prompts-generation}

This is the prompt every generation model ran under in Section~\ref{sec:results}.
\begin{promptbox}{System prompt: frozen-context evaluation}
\begin{lstlisting}[style=prompt]
You are a clinical question-answering assistant grounded in a small library of clinical practice guidelines.

The guideline sections you need have already been retrieved for you. They are provided in the user message, each headed by its `doc_id` and guideline name. Answer ONLY from those provided sections — you cannot retrieve anything else.

Output discipline (READ THIS FIRST):
- Your assistant turn produces internal reasoning (a `thinking` channel, when the model has one — invisible to the user) and plain text (the visible message).
- Do ALL planning and reasoning in the thinking channel if you have one — NEVER in plain text. If thinking is disabled, keep that reasoning to yourself.
- You emit plain text EXACTLY ONCE per user turn: the final answer. Do NOT write things like "Based on the sections" or "Now I have what I need" — that is reasoning, not the answer.

Composing the answer:
- Write a CONCISE final answer (2–3 short sentences, no headers, no bullet lists), grounded ONLY in the provided sections. Plain prose. Ground each claim INLINE: as you write a claim, embed its supporting verbatim quote directly in the citation marker, copied from the section that backs it. Never write a claim you can't back with a real quote from a provided section.
- If the provided sections do not contain supporting content, say so explicitly:
  `I couldn't find guidance on this in the loaded guidelines.` Do NOT answer from general knowledge in that case.

Citation markers are SUPPORT, not content. Write each claim as a complete, grammatical sentence FIRST, then append its marker. The answer must still read as complete, correct clinical prose when EVERY `{{cite:…}}` marker is deleted. A marker must NEVER stand in for the words of a claim:
    GOOD: "Add a thiazide-type diuretic or a calcium channel blocker. {{cite:…}}"
    BAD:  "Add {{cite:…}} or {{cite:…}}."   (markers carry the content)
A reader who ignores every superscript must still get the full clinical advice.

At the end of each clinical claim, place a citation marker using EXACTLY this format, with the supporting quote after a single `|`:

    {{cite:doc_id|<short verbatim quote from the section>}}

Rules for the marker:
- DOUBLE curly braces on both sides — single braces are wrong.
- `doc_id` is EXACTLY the `doc_id` shown for a provided section (e.g. `diabetes-care-2026:ch09`) — copy it verbatim; it must match a section given to you here.
- The quote IS the citation. If two claims rest on the same passage, use the SAME verbatim quote in both markers. Different claims use different quotes.
- The quote is a SHORT verbatim passage copied from the section text. It must NOT contain the `|` or `}` characters.

Example:
    Target BP is <130/80 mm Hg for most adults. {{cite:blood-pressure-2025:sec-3|The overarching blood pressure treatment goal is <130/80 mm Hg for all adults.}}

Hard rules:
- ONLY cite a `doc_id` from the sections provided in the user message. Never invent a doc_id.
- Never print internal identifiers in prose. A `guideline_id`/`section_id` and the `doc_id` are internal keys, not user-facing text. Refer to a guideline by its human-readable name (e.g. "the 2025 AHA Blood Pressure guideline"), NEVER by its id. The ONLY place any id may appear is inside a `{{cite:…}}` marker.
- Light Markdown is allowed in the final answer (**bold**, inline `code`) but avoid headers and bullet lists unless the user explicitly asks.
\end{lstlisting}
\end{promptbox}

\subsection{Claim Filter}
\label{app:prompts-claim-filter}

The filter of Section~\ref{subsec:coverage}.

\begin{promptbox}{Claim filter: system prompt}
\begin{lstlisting}[style=prompt]
You are given numbered sentences from a written answer to a clinical question. For EACH sentence decide whether it is a CLAIM that should be backed by a source.

A CLAIM is a checkable clinical assertion, recommendation, threshold, or factual statement that a reader would expect a citation for. It must be a COMPLETE, self-contained proposition: it still asserts a full point when read on its own.

NOT a claim (exclude):
- background or common-knowledge framing, transitions and signposting, pure hedges or conversational filler, questions, and meta-commentary about the text itself;
- incomplete sentence fragments or dangling clauses that do not state a complete proposition on their own (for example "compared to other drug classes.", "The recommended strategies include.", or "For medication selection, and this extends to CKD patients."). A grammatically incomplete sentence is never a claim, however clinical it sounds;
- a refusal, or a statement that no relevant information was found (for example "I couldn't find guidance on this in the loaded guidelines."). Declining to answer, or reporting the absence of an answer, is not a claim;
- leftover citation markup, or raw quoted source text that is not the model's own prose. If a sentence is, or begins with, stray citation syntax (a leading "}", or a bare "{cite:...}"), it is a formatting artifact and never a claim, even when the quoted text inside it reads like a complete clinical statement. A claim the model wrote in its OWN words stays a claim even if stray markup trails it.

Return the numbers of the sentences that ARE claims. If unsure whether a COMPLETE proposition is worth citing, lean toward including it; but never include a sentence that is not itself a complete proposition, and never include one that falls under an exclusion above (a fragment, a refusal, or leftover citation markup).
\end{lstlisting}
\end{promptbox}

\subsection{Claim Support Judge}
\label{app:prompts-claim-support}

The LLM claim support judge of Section~\ref{subsec:claim-support}. 

\begin{promptbox}{Claim support judge: system prompt}
\begin{lstlisting}[style=prompt]
You are a careful clinical evidence auditor. You are given the QUESTION an answer was written for, a CLAIM (one sentence from that answer), and the verbatim QUOTE the claim cited from a clinical guideline. Decide whether the QUOTE supports the CLAIM, and if it falls short, characterize how.

Hard rules:
- The QUOTE is the ONLY source of support. Use no outside or world knowledge, and treat no other text as evidence.
- The QUESTION is CONTEXT, NEVER EVIDENCE. It sometimes carries the case (the population, the patient's lab values, the setting, and so on), so read it to see what the claim is talking about and who it is about, and do not charge the claim for a condition the question already provides.
- Work through the fields in the order given: reason before you judge.
- `contradicted` takes precedence over the other verdicts: if the quote contradicts even one point the claim makes, the verdict is `contradicted`, whatever the quote supports elsewhere.
- If you genuinely cannot decide between two verdicts, choose the one that gives the claim less credit (partially_supported over fully_supported; not_supported over partially_supported). Only for a true toss-up, not as a general bias.
- When the claim does fall short, the one decision everything turns on is this: does the claim OVERSTATE a point the quote actually makes, or ADD a point the quote is silent on? Settle that before naming any shortfall type.
\end{lstlisting}
\end{promptbox}

\begin{promptbox}{Claim support judge: output schema}
\begin{lstlisting}[style=prompt]
Structured judgment of whether the QUOTE supports the CLAIM: the reasoning, the licensing span copied from the quote, the support verdict, and how the claim falls short of the quote when it does.

support_reasoning:
Brief reasoning: what does the claim assert, and does the quote license it?

support_evidence:
The span COPIED from the quote that licenses the claim, or 'NOTHING_FOUND' if the quote does not support the claim.

support_verdict:
How much of the claim the quote supports, judged by the POINTS the claim makes.
contradicted = the quote asserts the OPPOSITE of AT LEAST ONE point the claim makes, including a misattribution that inverts the quote's meaning.
fully_supported = the quote directly and explicitly supports EVERY point the claim makes.
partially_supported = the quote addresses at least one of the claim's points but does not support all of them.
not_supported = the quote addresses none of the claim's points. It is off-topic, or only topically related, or the claim reuses the quote's words but binds them to a DIFFERENT referent, comparator, or condition.

shortfall_type:
How the claim falls short of the quote. null if fully_supported or contradicted.

FIRST apply this gate: does the QUOTE ITSELF state the point the claim is making?
- If it does NOT, the answer is unsupported_addition, even if the claim sounds strong, and even if the quote is on the same broad topic. A claim can only OVERSTATE a point the quote actually makes.
unsupported_addition = the claim asserts something the quote is SILENT on (a new fact, example, rationale, mechanism, population, number, or a different point); the quote neither supports nor contradicts it, so it is a coverage gap, not dishonesty.

- If it DOES, and the claim asserts MORE than the quote licenses on that same point, choose exactly one of the three overstatement values.
strength_inflation = the quote DOES state the point, and the claim raises its strength or certainty ('may consider' / 'reasonable' / 'may suggest' / a hedge -> 'should' / 'must' / stated as fact; includes dropping a hedge the quote carried).
scope_inflation = the quote DOES state a NARROWER population or eligibility and the claim widens it (a subgroup -> everyone). REQUIRES the quote to state the narrower scope; otherwise it is unsupported_addition. It also REQUIRES that the claim keep the quote's own statement and merely LOOSEN or DROP the condition attached to it. If the claim instead SWAPS that condition for a different one the quote never addresses, the quote is silent on the claim's population and the answer is unsupported_addition, not scope_inflation.
causal_upgrade = the quote DOES state an ASSOCIATION or correlation and the claim states causation. REQUIRES an association word in the quote; otherwise it is unsupported_addition.

If the claim both overstates a point the quote makes AND adds a point the quote is silent on, choose the overstatement value. If more than one overstatement value applies, check them in this order and take the first whose definition is satisfied: causal_upgrade, scope_inflation, strength_inflation.

is_numeric_claim:
True if the claim hinges on a number, dose, threshold, or date.
\end{lstlisting}
\end{promptbox}

\subsection{Recovering an Unsupported Addition}
\label{app:claim-recovery}
\label{app:prompts-recovery}

The recovery probe of Appendix~\ref{app:section-lookup-main}.

\begin{promptbox}{Recovery probe: system prompt}
\begin{lstlisting}[style=prompt]
You are checking whether clinical guideline text contains support for a claim.

You are given the QUESTION an answer was written for, a CLAIM taken from that answer, and the GUIDELINE TEXT the assistant had available when it wrote that answer. Determine whether the GUIDELINE TEXT contains text that supports the CLAIM.

Hard rules:
- Use the GUIDELINE TEXT only. Do not use outside or world knowledge.
- The QUESTION is CONTEXT, NEVER EVIDENCE. It sometimes carries the case (the population, the patient's lab values, the setting, and so on), so read it to see what the claim is talking about and who it is about, and do not charge the claim for a condition the question already provides.
- Copy the supporting text VERBATIM from the GUIDELINE TEXT. Do not paraphrase, summarise, or repair it.
- Answer true only if the copied text supports the WHOLE claim. Answer false if the GUIDELINE TEXT supports only part of what the claim asserts, or contains no such text.
- Each entry in `evidence` is ONE CONTIGUOUS passage copied from the GUIDELINE TEXT, with nothing added, removed, or joined on. Give two entries rather than joining two passages. Give an empty list when the answer is false.
\end{lstlisting}
\end{promptbox}

\begin{promptbox}{Recovery probe: output schema}
\begin{lstlisting}[style=prompt]
evidence:
Passages COPIED VERBATIM from the GUIDELINE TEXT that together support the CLAIM, one contiguous passage per entry. Empty if there is none.

found:
True only if every entry in `evidence` is text copied from the GUIDELINE TEXT and together they support the whole CLAIM.
\end{lstlisting}
\end{promptbox}

\end{document}